\documentclass[a4paper,11pt]{elsarticle}
\usepackage{graphicx}		% include graphics
\usepackage{url}
\usepackage[utf8]{inputenc}  	% for czech diacritics (used mainly in names)
\usepackage{subcaption}
\usepackage{multirow}
\usepackage{amsmath}
\usepackage{natbib}

\usepackage[usenames,dvipsnames]{color}

\makeatletter
\def\convertto#1#2{\strip@pt\dimexpr #2*65536/\number\dimexpr 1#1}
\makeatother

\journal{Neurocomputing}

\begin{document}
\begin{frontmatter}
\date{}

\title{Self-supervised network distillation: an effective approach to exploration in sparse reward environments}

\author[1,*]{Matej Pech\'a\v{c}}
\author[2]{Michal Chovanec}
\author[1]{Igor Farka\v{s}}

\address[1]{Department of Applied Informatics, Comenius University Bratislava, Slovak Republic}
\address[2]{Excalibur, Ltd., Poprad, Slovak Republic}
\address[*]{Correspondence: matej.pechac@fmph.uniba.sk}

%\ead{\{matej.pechac,igor.farkas\}@fmph.uniba.sk, mchovanec@photoneo.com}
%Photoneo Ltd. \\ 

%%%%%%%%%%%%%%%%%%%%%%%%%%%%%%%%%%%% 
\begin{abstract}
Reinforcement learning can solve decision-making problems and train an agent to behave in an environment according to a predesigned reward function. However, such an approach becomes very problematic if the reward is too sparse and so the agent does not come across the reward during the environmental exploration. The solution to such a problem may be to equip the agent with an intrinsic motivation that will provide informed exploration during which the agent is likely to also encounter external reward. Novelty detection is one of the promising branches of intrinsic motivation research. We present Self-supervised Network Distillation (SND), a class of intrinsic motivation algorithms based on the distillation error as a novelty indicator, where the predictor model and the target model are both trained. We adapted three existing self-supervised methods for this purpose and experimentally tested them on a set of ten environments that are considered difficult to explore. The results show that our approach achieves faster growth and higher external reward for the same training time compared to the baseline models, which implies improved exploration in a very sparse reward environment.\footnote{The source code is available at \url{https://github.com/Iskandor/SND} and \url{https://github.com/michalnand/reinforcement_learning}. The video of our trained agents is available at \url{https://youtu.be/-vDg_r2ZetI}.}
In addition, the analytical methods we applied provide valuable explanatory insights into our proposed models.
\end{abstract}

\end{frontmatter}

%%%%%%%%%%%%%%%%%%%%%%%%%%%%%%%%%%%%%%%%%%%%%%%%%%%%%%%%%%%%%%%%%%%%%%% INTRO
\section{Introduction}
\label{sec:intro}

The development of reinforcement learning (RL) methods has achieved much success over the last decade, since together with advances in computer vision  \citep{krizhevsky2012imagenet,he2016deep}, it became possible to teach agents to solve various tasks, play computer games \citep{mnih2013playing} (see overview in \citep{RLgames2023}) even surpassing human players \citep{mnih2015humanlevel}.
Nevertheless, these single tasks require very long training times and a lot of computational resources.
Coping with complex environments such as the real world is still a challenge. There are several research opportunities, one of them being the search for more efficient learning methods.

The complex environments with sparse rewards pose a special challenge for RL approaches.
The most popular computational approach to make RL more efficient is based on a concept of {\it intrinsic motivation} (IM) \citep{baldassarre2014intrinsic}. 
IM has a strong biological basis  \citep{Ryan00,Morris2022}, since it is observed among higher animals, especially in humans, engaging them in various activities. 
Intrinsic motivations appear early after birth and guide the biological agents during their entire lives. IM is considered one of the prerequisites for open-ended (or, life-long) learning. 
If we want to achieve this capacity with artificial agents \citep{Parisi2019}, we have to master this first step and equip them with an ability to generate their own goals and acquire new skills \cite{Holas2021}. 
Therefore, computational approaches concerned with IMs and open-ended development provide the potential in this direction leading to more intelligent systems, in particular those capable of improving their own skills and knowledge autonomously and indefinitely \citep{baldassarre2014intrinsic,Baldassarre19}.

The concept of intrinsic (and extrinsic) motivation was first studied in psychology \citep{Ryan00} and later it entered the RL literature \citep{barto2005intrinsic, singh2010intrinsically, Barto2013}. The first taxonomy of computational models appeared in \cite{oudeyer2009intrinsic} where the concept of motivation is divided into external and internal, depending on a mechanism that generates motivation for the agent. \textit{External} motivation assumes the source of motivation coming from outside the agent, and it is always associated with a particular goal in the environment. If the motivation is generated within the structures that make up the agent, this implies an \textit{internal} motivation. 

Another dimension for the differentiation, extrinsic or intrinsic, is less obvious (see also \citep{Morris2022}).
\textit{Extrinsic} motivations pertain to behaviors whenever an activity is performed in order to attain some separable outcome.\footnote{For instance, a child studies diligently at school not just enjoying the activities as such, but for a certain reason, e.g. expecting a better job in the future.}
Some variability exists in this context since these behaviors can vary in the extent to which they represent self-determination (see the details in \citep{Ryan00}).
On the other hand, \textit{intrinsic} motivation is defined as doing an activity for its inherent satisfaction. IM has been operationally defined in various ways, backed up by different psychological theories which point to some uncertainty in what it exactly means. Nevertheless,
a solution of an operational definition of IMs has been proposed \cite{Baldassarre19} in terms of processes that can drive the acquisition of knowledge and skills in the absence of extrinsic motivations. Furthermore, the author proposes (and explains why) a new term of \textit{epistemic motivations} as a suitable substitution for intrinsic motivations. 
Despite some uncertainty, intrinsic motivation has remained a well-coined term in the literature.

Intrinsic motivation is a crucial factor that helps the agent not only to remain in open-ended learning, hence solving a variety of tasks \citep{Parisi2019}, but it also helps to solve a single task with extremely sparse rewards. In this paper, we focus on this case.

There exists a variety of approaches aiming to use IM-based signals for agent learning. Information-theoretic view on IM is well represented in the literature,  involving the concepts of novelty, surprise, and skill learning. The recent review \citep{Aubret2023} suggests that novelty and surprise can assist the building of a hierarchy of transferable skills that abstract dynamics and make the exploration process more robust. In this context, abstraction is a key feature of the agent architecture where it makes sense to introduce learning mechanisms to enforce the formation of proper internal representations that lead to improved agent performance. 

Learning appropriate internal representations from unlabelled input data (e.g. images) for the purpose of solving various problems is in general a useful task in machine learning. This can be achieved in various ways (related methods are mentioned in Sec.~\ref{sec:related-work}), including supervised end-to-end (deep) learning, self-supervised autoencoders, unsupervised feature extractors (such as contrastive divergence learning) and RL-based approaches where a certain loss function is optimized.

As our main contribution, we identified shortcomings of the Random Network Distillation model (RND) \cite{burda2018exploration} that served as a starting point for our work. RND is a novelty detection method that uses two models, the target model and the predictor model, both being fed by the current state as their input. The predictor model attempts to imitate the target model, and their difference (in terms of generated features) is used for intrinsic motivation. 
However, the RND method requires good initialization of the target model, has a low variance of intrinsic rewards, and loses the motivational signal, which is caused by the adaptation of the predictor model.

\begin{figure*}[t!]
\centering
\includegraphics[width=0.8\textwidth]{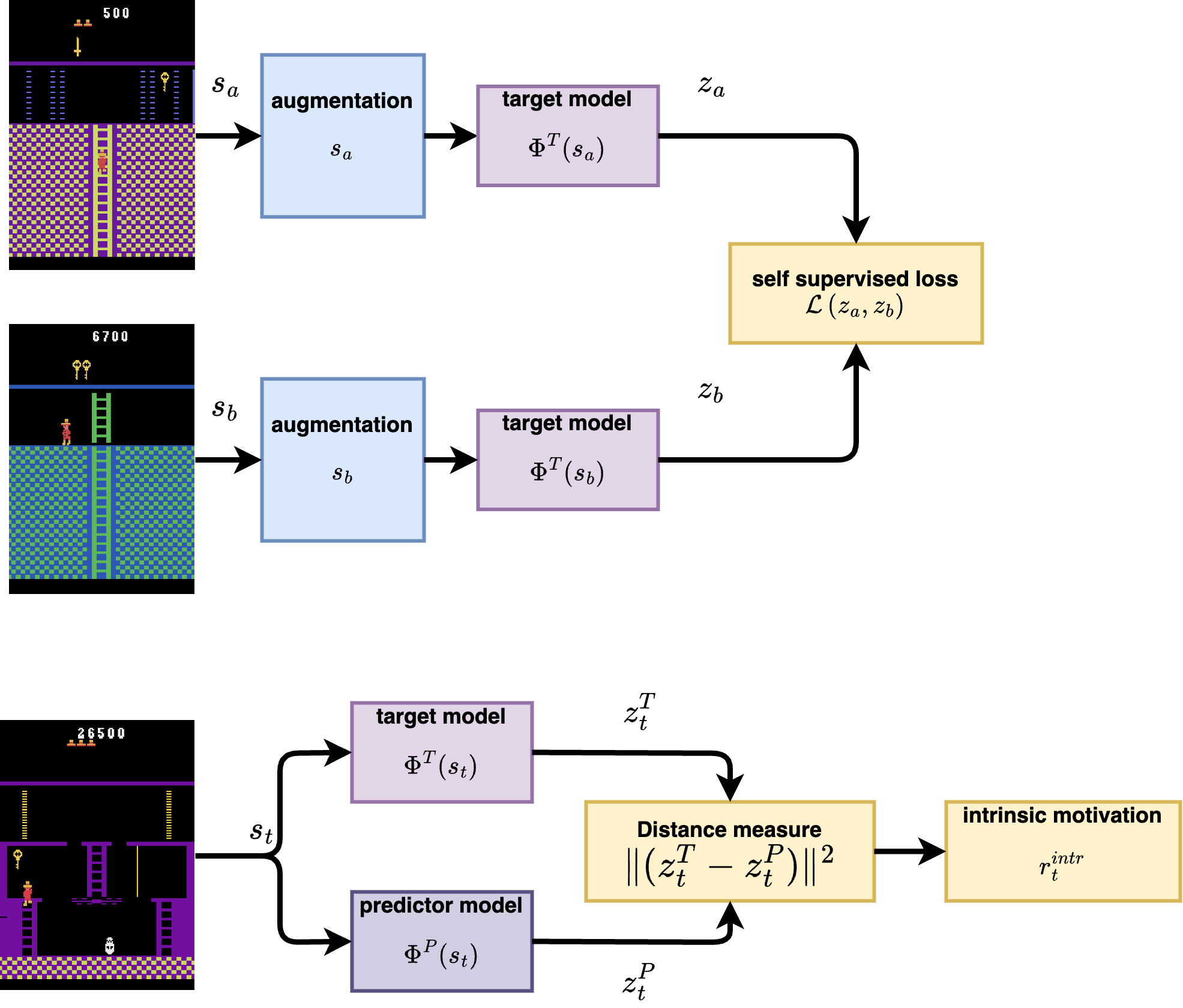}
\caption{Self-supervised network distillation (SND) principle. The proposed method consists of two main parts. {\it Top:} Self-supervised learning of the suitable features for the target model. {\it Bottom:} Calculation of the intrinsic reward by target model distillation, using the squared Euclidean distance between the models' outputs.}
\label{fig:cnd_overview}
\end{figure*}

To address these shortcomings, in this contribution, we introduce a class of motivational methods based on the exploitation of the distillation error as novelty detection. 
To provide features with higher variance and better sensitivity for novelty detection, we use self-supervised regularization of the target model. Our methods were able to use more feature space dimensions compared to RND (which has an impact on the size of the distillation error) and more importantly, the distances between representations of very similar states increased, which achieved a better ability to distinguish them from each other and thus the task for the predictor model became more complex. This led to an increase in the variance of the intrinsic reward and a slowdown in the adaptation of the predictor model.
In particular, we adapted three existing self-supervised methods. 
The developed methodologies were tested on a set of environments that are considered difficult to explore.
The overall concept is shown in Figure~\ref{fig:cnd_overview}.

Using these methods, we were able to solve hard exploration seeds for ProcGen and the complete first level of the infamous Atari game -- Montezuma's Revenge, which, as far as we know, has never been finished using internal motivation methods. Overall, our models achieved excellent performance in most environments, and we provide insights based on our analyses, why this could be the case.

Our research goal was not to solve a general problem of RL, but as in many similar papers (see the related work) we focused on a set of challenging environments of a certain type (complex Atari games) with naturally sparse rewards. We also add challenging ProcGen games, since their authors provided hard exploration seeds, making the PPO algorithm unable to solve the tasks, even with a larger IMPALA model \citep{Espeholt2018}.

The rest of the paper is organized as follows. Section~\ref{sec:related-work} relates our work in a wider and narrower context. Section~\ref{sec:methods} describes the methods that we introduced and tested in this work. Section~\ref{sec:exper} explains in detail all experiments performed. Section~\ref{sec:discussion}  provides the discussion about the achieved results, the limitations of our models, and hints for further extensions. 
Section~\ref{sec:conclusion} concludes the paper by summarizing the main take-home messages.

%%%%%%%%%%%%%%%%%%%%%%%%%%%%%%%%%%%%%%%%%%%%%%%%%%%%%%%%%%%%%%%%%%%%%% CHAPTER 2
\section{Related work}
\label{sec:related-work}

We first provide the wider conceptual context to our work, and then focus more on methods that are closely related to our models.
According to the prevailing view, the approaches to IM can be divided into two main categories with adaptive motivations.
\textit{Knowledge-based} approach is focused on the acquisition of the world knowledge of and it draws on the theory of drives \citep{hull1943principles}, theory of cognitive dissonance \citep{festinger1962theory} and optimal incongruity theory \citep{hunt1965intrinsic}. \textit{Competence-based} approach focuses on the acquisition of skills by motivating the agent to achieve a higher level of performance in the environment, which means acquiring desired actions to achieve self-generated goals. Its psychological basis includes the theory of effectance \citep{white1959motivation} and the theory of flow \citep{Csikszentmihalyi91}. 

The knowledge-based category focusing on exploration can be divided into \textit{prediction-based}, \textit{novelty-based} and \textit{information-based} approaches \citep{aubret2019survey}.
Prediction-based approaches use the prediction error as an intrinsic reward signal. The source of error can be a forward model (e.g.~\cite{stadie2015incentivizing,bellemare13arcade,Pathak2017}), a generative model \cite{yu2020intrinsic} (e.g.~based on a variational auto-encoder \cite{kingma2013auto}) or disagreement in a learned world model \citep{sekar2020planning}. Exploration with Mutual Information (EMI) \citep{kim2018emi} extracts predictive signals that can be used to guide exploration based on forward prediction in the representation space. Model-based Active eXploration (MAX) \citep{shyam2019model} uses an ensemble of forward models to plan to observe novel events.

Information approaches use quantities from information theory \citep{shannon1948mathematical}, such as information gain, mutual information, and entropy and try to maximize the information obtained by the agent from the environment. Variational Information Maximizing Exploration (VIME) \citep{houthooft2016vime} approximates the environment dynamics and uses the information gain of the learned dynamics model as an intrinsic reward. Random Encoders for Efficient Exploration (RE3) \citep{seo2021state} is an exploration method that utilizes state entropy as an intrinsic reward.
For more details, we recommend surveys \cite{burda2018large, aubret2019survey, yuan2022intrinsically}.

In a narrower context, our work falls into the area of novelty-based methods, which are part of knoweldge-based approach, that monitor state novelty.
The first models exploited the count-based approach \citep{tang2017exploration}. This approach is impractical for large or continuous state spaces and it was extended by introducing pseudo-count and neural density models \citep{ostrovski2017count,martin2017count,machado2018count}. A similar method to pseudo-count was used by a random network distillation (RND) model \citep{burda2018exploration} with a lower complexity. RND method served as the basis for our SND methods and finally, we showed, that SND is a generalized case of RND. The never-give-up framework \citep{badia2020never} learns intrinsic rewards composed of episodic and life-long state novelty (which is detected by an RND model). BYOL-Explore \citep{guo2022byol} is one of the recent methods using self-supervised learning in the context of exploration and intrinsic rewards. The method achieved outstanding results in hard exploration environments which we consider to support our thesis that self-supervised learning algorithms can significantly improve intrinsic motivation methods when used correctly. Therefore, we consider BYOL-Explore to be conceptually similar to SND.

Self-supervised learning is a paradigm of machine learning when the agent does not have any externally labeled data. The goal is to use the information in the data itself and prepare the model to perform another task. Self-supervised learning also started to be used in the field of state representation learning \citep{Timoth2018}, it has proven to be a suitable method for creating the feature space \citep{Anand2019} and has also found its use in reinforcement learning \citep{Srinivas2020, guo2022byol}. Contrastive learning \citep{Chopra2005} is a method of self-supervised learning minimizing a discriminative loss function that drives the similarity metric to be small for pairs from the same category, and large for pairs from different categories.
Several different objective functions have been proposed, e.g. Noise Contrastive Estimation (NCE) \citep{Gutmann2010}, InfoNCE \citep{Oord2018}, or multi-class $N$-pair loss \citep{Sohn2016}. Another approach to self-supervised learning is based on regularization (non-contrastive methods). In this case, the model sees only positive examples (data points from the same category) and generates a feature space based on various regularization losses like invariance and covariance loss in the case of Barlow Twins model \citep{Zbontar2021}, triple variance-invariance-covariance losses in VICReg model \citep{Bardes2022}, or other priors like proportionality, variability, slowness principle, or repeatability \citep{jonschkowski2015learning}.
Bootstrap Your Own Latent \citep{grill2020bootstrap} relies on two neural networks, referred to as online and target networks that interact and learn from each other. This self-supervised method is the core idea of the intrinsic motivation method BYOL-Explore.

%%%%%%%%%%%%%%%%%%%%%%%%%%%%%%%%%%%%%%%%%%%%%%%%%%%%%%%%%%%%%%%%%%%%%%%% CHAPTER 3
\section{Methods}
\label{sec:methods}

The decision-making problem in the environment using RL is formalized as a Markov decision process defined as a 5-tuple ($\mathcal{S,A,T,R}, \gamma$)
consisting of a state space $\mathcal{S}$ with actions $s_t$, action space $\mathcal{A}$ with actions $a_t$, transition function $\mathcal{T}_a(s,s') = P(s_{t+1} = s'|s_t = s, a_{t} = a)$, reward function $\mathcal{R}_a(s,s')$ specifying the immediate (external) reward after transition from $s$ to $s'$ with action $a$, and a discount factor $\gamma$. The main goal of the agent is to maximize the discounted return  
$R_t = \sum_{k=0}^\infty \gamma^k r_{t+k}$
in each state, where $r_t$ is an immediate external reward at time $t$.
The stochastic policy is defined as a state-dependent probability function $\pi : \mathcal{S}\times\mathcal{A} \rightarrow [0, 1]$ such that
$\pi(s,a) = P(a_t = a | s_t = s)$ and $\sum_{a \in \mathcal{A}} \pi_{t}(s,a) = 1$,
and the deterministic policy is a mapping $\pi: \mathcal{S}\rightarrow \mathcal{A}$.

The methods searching for the optimal policy $\pi^{*}$, that maximizes the expected return $R$, can be divided into on-policy (e.g. SARSA \citep{rummery1994line}, A2C \citep{mnih2016asynchronous}, PPO \cite{schulman2017proximal})
and off-policy methods (the family of Q-learning algorithms) \cite{mnih2013playing}.
The actor--critic algorithms are based on two separate modules: an \textit{actor} generates actions following the agent's policy $\pi$ and a \textit{critic} estimates the state value function $V^{\pi}$ defined for a stochastic policy as\footnote{With a slight abuse of mathematical notation, for a deterministic policy, $\pi(s,a)=1$ if $\pi(s)=a$, otherwise $\pi(s,a)=0$.}
$$
V^{\pi}(s) = \sum_a \pi(s,a) \sum_{s'} \mathcal{T}_{s,a,s'} \left[ \mathcal{R}_{s,a,s'} + \gamma V^{\pi}(s') \right]
$$
or the state-action value function $Q^{\pi}$ defined as
$$
Q^{\pi}(s,a) = \sum_{s'} \mathcal{T}_{s,a,s'} \left[ \mathcal{R}_{s,a,s'} + \gamma V^{\pi}(s') \right].
$$
The actor then updates its policy to maximize return $R$ based on the critic's value function estimations.

In high-dimensional tasks, when one cannot use the Bellman equation, the common approach is to use function approximators (e.g., deep neural networks) for estimating the critic and the actor functions.

\subsection{Intrinsic motivation and exploration}

During the learning process, the agent must explore the environment to encounter an external reward and learn to maximize it. 
In the case of a deterministic policy, noise needs to be added to the state-conditional action selection. 
In both stochastic and deterministic cases, these are {\it uninformed} environmental exploration strategies. The problem arises if the external reward is very sparse and the agent cannot use these strategies to find the sources of the reward. In such a case, it is advantageous to use {\it informed} strategies which include the concept of intrinsic motivation.

In the context of RL, intrinsic motivation can be realized in various ways, but most often it is a new reward signal $r^{\rm intr}_t$ scaled by the parameter $\eta$ which is generated by what we refer to as the motivational module and is added to the external reward $r^{\rm ext}_t$
\begin{equation}
\label{eq:rintr}
  r_t = r_t^{\rm ext} + \eta \, r_t^{\rm intr}.
\end{equation}
The goal of introducing such a new reward signal is to provide the agent with a source of information that is absent from the environment when the reward is sparse, and thus facilitate the exploration of the environment and the search for an external reward.

\begin{figure*}[thb]
  \centering
  \includegraphics[width=10cm]{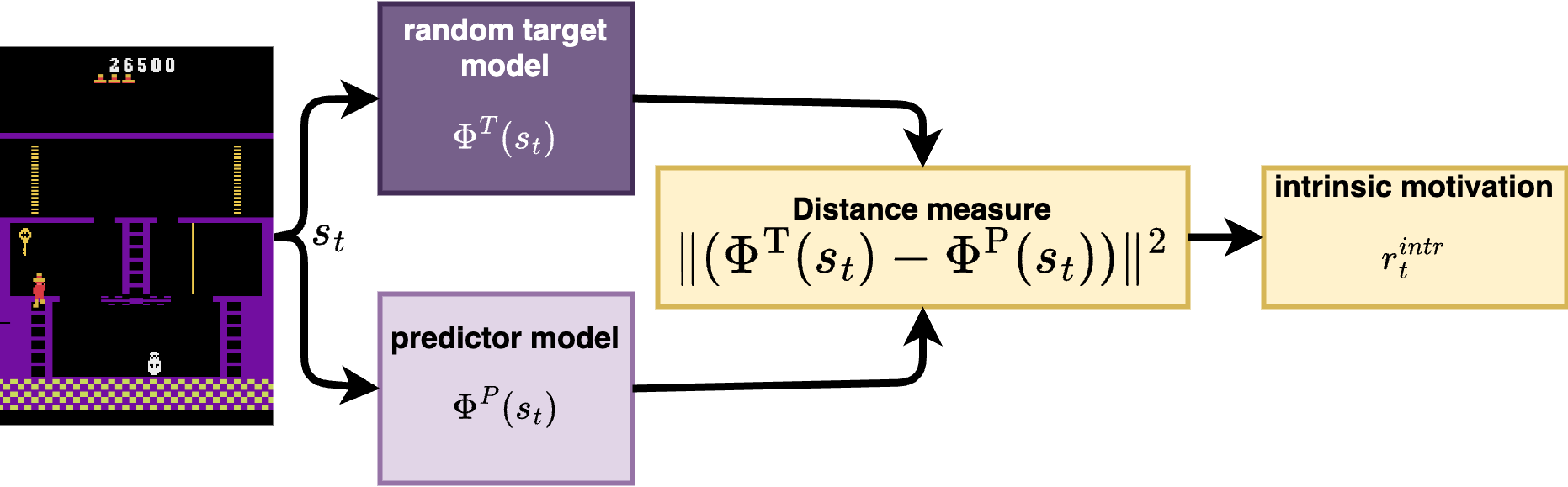}
  \caption{The basic principle of generating an exploration signal in Random Network Distillation.}
\label{fig:cnd_rnd}
\end{figure*}

\subsection{Random Network Distillation}
The Random Network Distillation (RND) \citep{burda2018exploration} model shown in Figure~\ref{fig:cnd_rnd} is a representative of methods based on novelty detection, belonging to the knowledge-based category of IM. The motivation module has two components: the target model $\Phi^{\rm T}$ that generates state representations and the predictor model $\Phi^{\rm P}$ that learns to estimate them. This process is called knowledge distillation. Intrinsic motivation, expressed as an intrinsic reward, is computed as the distillation error
\begin{equation}
\label{eq:distill_error}
r_{t}^{\rm intr} = \| (\Phi^{\rm T}(s_t) - \Phi^{\rm P}(s_t)) \|_2^2
\end{equation}
where $\|.\|^2_2$ denotes the squared $L_2$ vector norm.
It is assumed that the predictor network $\Phi^{\rm P}$ will be able to more easily estimate the feature vectors for states it has seen multiple times, while new or rarely visited states will induce a large distillation error.
The approach is simple and successful in environments with sparse reward but has two serious drawbacks: (1) It is necessary to properly initialize the random network and (2) over time, the signal of intrinsic motivation disappears due to sufficient adaptation of the learning network (a phenomenon that could be called generalization). In the section dedicated to analyzing the results of our method, we show that the RND method can detect only large changes in the state space, while small changes are almost invisible to it. 
This is a consequence of inappropriate features provided by the target model.

\subsection{Self-supervised Network Distillation}

We modified the concept of distillation of randomly initialized static network RND \citep{burda2018exploration} and instead, we distilled a network that learns continuously using a self-supervised algorithm. We denote the methods SND.

Let $X = \{(s_t, s_{t+1})_n\}_{n=1}^N$ be a minibatch of $N$ consecutive states from $\mathcal{S}$ that are sampled from collected trajectories and $\mathcal{Z}$ is a $D$-dimensional latent space of state representations. Let $S$ correspond to the set of actual states and $S_{\rm next}$ correspond to the set of next states (i.e. the set of left-column states and right-column states in each pair of states respectively). The shape of states $S$ and $S_{\rm next}$ in our experiments is a batch of stacked frames: $(N, C, H, W)$, where $N$ is the batch size, $C$ is the number of stacked frames, $H$ and $W$ are the height and the width of a frame. 
The architecture of the SND motivation module in general consists of a target model $\Phi^{\rm T}: \mathcal{S} \rightarrow \mathcal{Z}$ and a predictor model $\Phi^{\rm P}: \mathcal{S} \rightarrow \mathcal{Z}$ but with an {\it essential difference} that the network generating the target feature vectors (target model) {\it also learns}. The corresponding output $Z = \Phi^{\rm T}(S)$ is a $N$$\times$$D$ matrix (so the feature vectors are in rows). 
The predictor model optimizes the loss function
\begin{equation}
\label{eq:predictor_loss}
\mathcal{L}_{P} = \frac{1}{N} \sum_{s \in S} \| (\Phi^{\rm T}(s) - \Phi^{\rm P}(s)) \|_2^2
\end{equation}
The target model optimizes the loss function $\mathcal{L}_{T}$. Further, we introduce three methods (loss functions) for training the target model based on different self-supervised algorithms.

\begin{figure*}[ht]
\centering
\includegraphics[width=12cm]{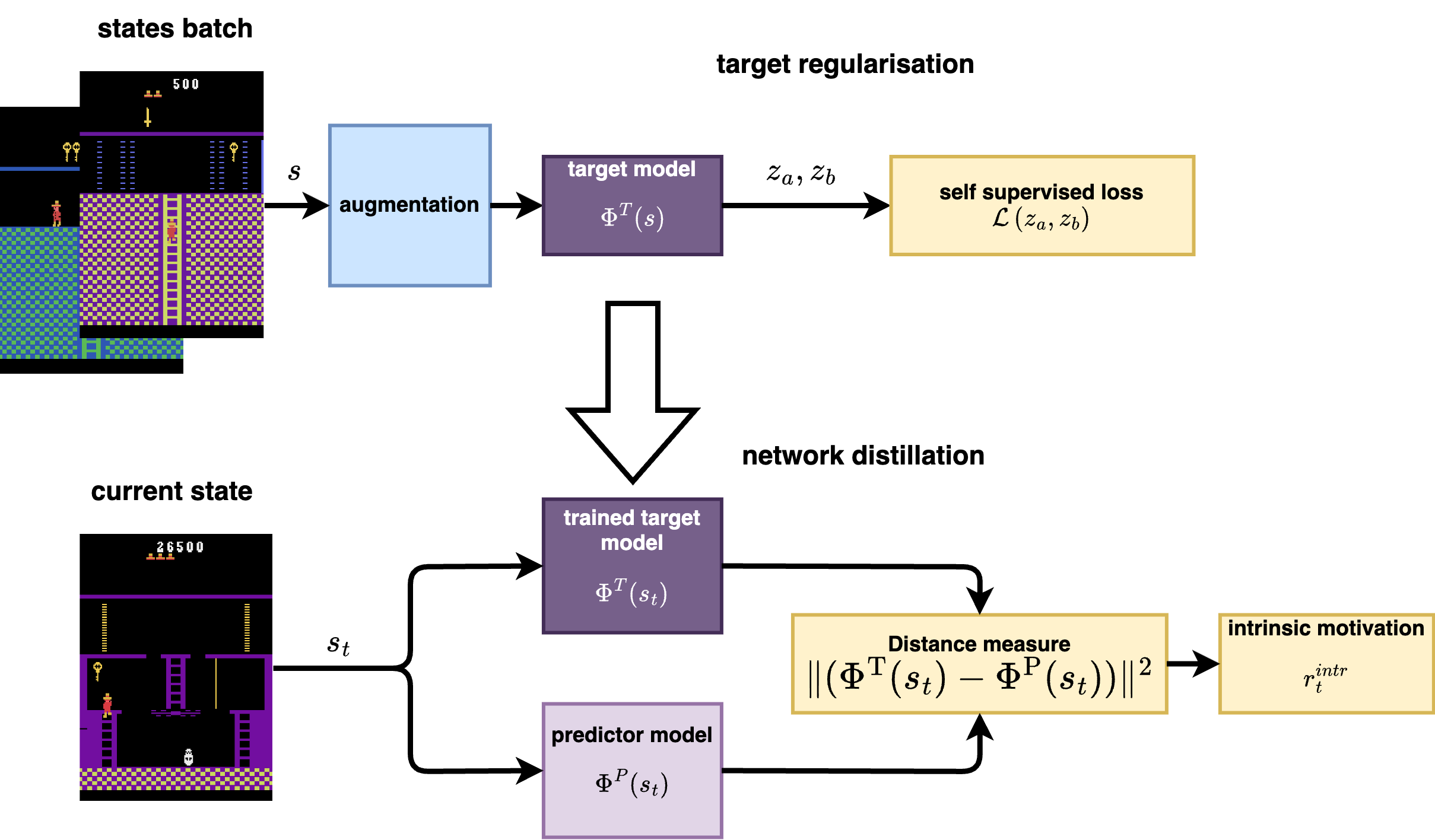}
  \caption{The basic principle of generating an exploration signal in Self-supervised Network Distillation. The calculation of the intrinsic reward is the same as for RND, but in these methods, the target model changes over time and generates a more complex feature space.}
  \label{fig:cnd_cnd}
\end{figure*}

Both models are trained in the same loop as the actor and the critic. 
With a trained target model, the task for the predictor model becomes much more difficult, since the representations of the states move further apart and their distribution in the feature space becomes very different from the distribution of the states in the state space. This leads to the elimination of the shortcomings associated with the RND method. These properties of the SND methods are described in detail in section~\ref{sec:analysis} devoted to the analysis of results. 
The schematic representation of the proposed approach is shown in Figure~\ref{fig:cnd_cnd}.

\begin{figure*}[thb]
\centering
\includegraphics[width=8cm]{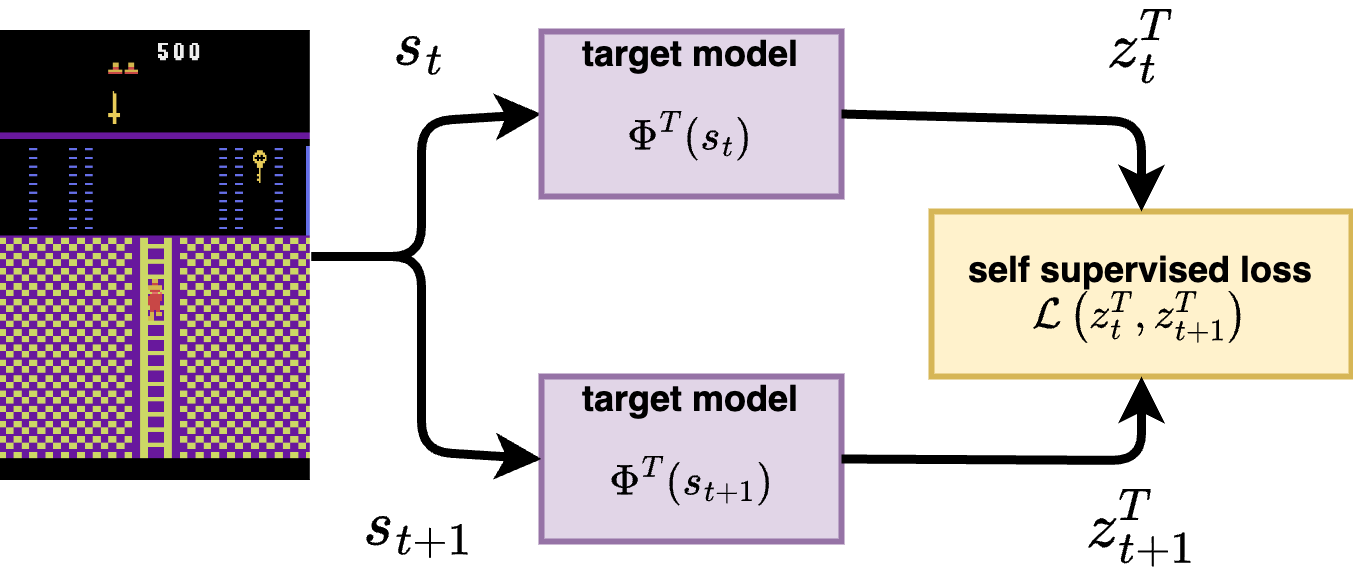}
\caption{Training of the SND target model using two consecutive states and the self-supervised learning algorithm.}
\label{fig:std_dim_idea}
\end{figure*}

%\MC{je to tak, a podla mna dosiahnutie metrickej vlastnosti tychto features bude klucove pre RL - aby aspon priblizne platili trojuholnikove nerovnosti - nepriamo sa to da dosihanut ze sa povie, ze "rovnake" stavy su aj tie, ktore su napr. max 4 stepy od seba - tak to robim teraz, s paradnymi vysledkami}

\subsubsection{SND-V}
This method uses a contrastive loss function \citep{Chopra2005}. 
Assume a batch of states $(S, S')$ randomly sampled from the trajectory buffer. For each pair of states $(s_n, s'_n)$ a target distance is generated.
With probability $p=0.5$, $s_n = s'_n$, in which case the target distance is set to $\tau_n = 0$, otherwise $(s_n \neq s'_n)$ the target distance is set to $\tau_n = 1$. 
The distance of predicted target model features $\phi^{\rm T}(s_n)$ and $\phi^{\rm T}(s'_n)$ is compared with the target distance in the mean squared error (MSE) loss.

\begin{figure*}
\centering
\includegraphics[width=12cm]{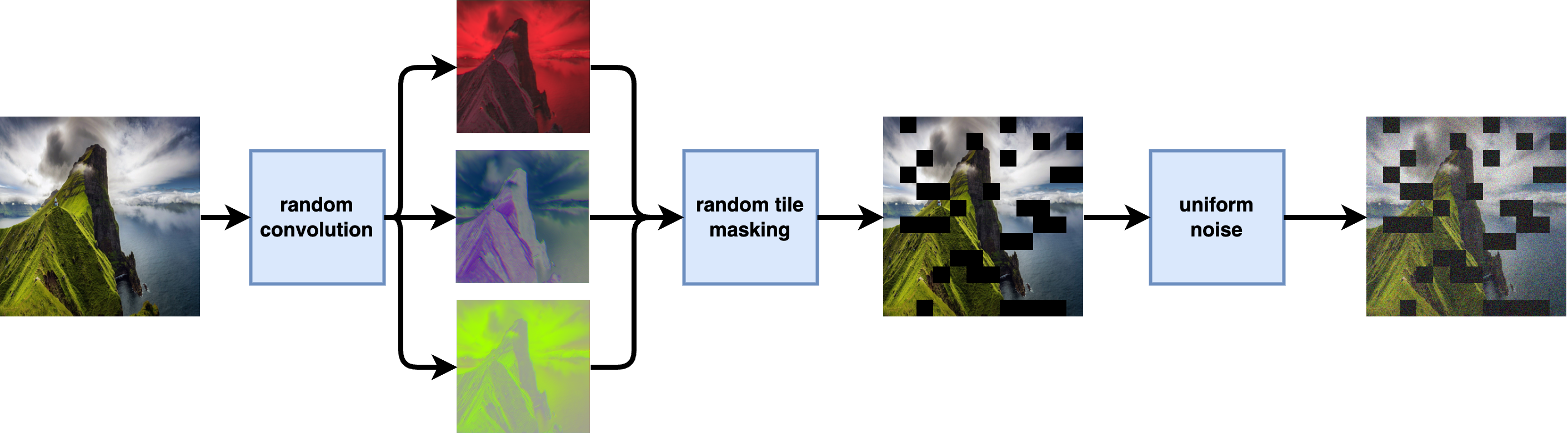}
\caption{The scheme of the state augmentation pipeline.}
\label{fig:stdv_augmentations}
\end{figure*}

For state preprocessing, we experimented with three augmentation schemes to learn suitable robust features:
\begin{enumerate}
  \item uniform noise only, from the range $\langle -0.2, 0.2\rangle$
  \item random tile masking + uniform noise, tile sizes 2, 4, 8, 12, 16
  \item random convolution filter + random tile masking + uniform noise.
\end{enumerate}
Uniform noise was used for each pixel value, the remaining augmentations with $p=0.5$. The whole pipeline is shown in Figure~\ref{fig:stdv_augmentations}. 
The idea of tiles masking can be supported by recently proposed self-supervised loss \citep{assran2022maskednetworks}. Random convolution was successfully used for a PPO agent in Procgen environment \citep{lee2020randomization}. And finally, adding noise is a common augmentation process.
The SND-V regularisation loss is defined as
\begin{equation}
\label{eq:sndv1}
\mathcal{L}_{T} = \sum_{n}(\tau_n - \|Z_n - Z'_n\|^2_2)^2
\end{equation}
where $\|.\|^2_2$ is the squared Euclidean distance between $n$-{\rm th} feature vectors $Z_n$ and $Z'_n$.
We also experimented with the loss function defined as
\begin{equation}
\label{eq:sndv2}
\mathcal{L}_{T} = 
  \begin{cases}
    \|Z_n - Z'_n\|^2_2 & \text{if $\tau_n = 0$} \\
    \max(1 - \|Z_n - Z'_n\|^2_2; 0) & \text{otherwise.}
  \end{cases}
\end{equation}
This loss function pulls the feature vectors of similar states (when $\tau_n = 0$) to one another by penalizing their squared distance. And on the contrary, it pushes away the feature vectors from one another, up to a certain limit when their distance exceeds one. We observed no improvements compared to Eq.~\ref{eq:sndv1}. This loss can be unstable because the max term omits the penalty for distances larger than one.

\subsubsection{SND-STD} 

This method uses the Spatio-Temporal DeepInfoMax (ST-DIM) algorithm \citep{Anand2019}.
The basic idea of the ST-DIM algorithm (see Figure~\ref{fig:std_dim_idea}) is to move closer the representations of two consecutive states on the output layer and the individual representations on the intermediate convolution layer that carry information about the location (global-local objective) and also to move closer the representations on the intermediate convolution layer for consecutive states (local-local objective).

In our context, the positive samples correspond to pairs of consecutive observations $(s_t, s_{t+1})$ and negative samples correspond to pairs of non-consecutive observations $(s_t, s^{*}_t)$, where $s^{*}_t$ is a
randomly sampled observation from the same minibatch. Hence, there are many more negative samples than positive samples.

Mathematically, ST-DIM uses two loss functions: 
the global-local objective (GL) and the local-local objective (LL), defined as
\citep{Sohn2016}: 
\begin{equation}
\label{eq:sndstd1}
\mathcal{L}_{\rm GL} = - \sum_{h=1}^{H} \sum_{w=1}^{W} \log \frac{ \exp (g_{h,w}(s_t,s_{t+1}))} { \sum_{s_{t}^{*} \in S_{\rm next}} \exp (g_{h,w}(s_t,s_t^{*}))}
\end{equation}
\begin{equation}
\label{eq:sndstd2}
\mathcal{L}_{\rm LL} = -  \sum_{h=1}^{H} \sum_{w=1}^{W} \log \frac{ \exp (f_{h,w}(s_t,s_{t+1}))} { \sum_{s_t^{*} \in S_{\rm next}} \exp (f_{h,w}(s_t,s_t^{*}))} 
\end{equation}
where the score function $g_{h,w}(s_t,s_{t+1}) = \Phi^{\rm T}(s_t) W_g \Phi^{\rm T}_{(l,h,w)}(s_{t+1})^\top$ is defined as the unnormalized cosine similarity between transformed global features $\Phi^{\rm T}(s_t)$ and local features $\Phi^{\rm T}_{(l,h,w)}(s_{t+1})$ of the intermediate convolution layer $l$ in $\Phi^{\rm T}$. Global features are transformed by (trained) matrix $W_g$ to the space of local features. Analogically, $f_{h,w}(s_t,s_{t+1}) = \Phi^{\rm T}_{(l,h,w)}(s_t) W_l \Phi^{\rm T}_{(l,h,w)}(s_{t+1})^\top$ is the unnormalized cosine similarity between transformed local features $\Phi^{\rm T}_{(l,h,w)}(s_t)$, by (trained) matrix $W_l$, and $\Phi^{\rm T}_{(l,h,w)}(s_{t+1})$. The resulting loss function is then defined as
\begin{equation}
\label{eq:sndstd3}
\mathcal{L} = \frac{1}{MN} (\mathcal{L}_{\rm GL} + \mathcal{L}_{\rm LL}).
\end{equation}
Following this objective function, the target model becomes a good feature extractor adapting to new states discovered by the agent. However, after initial tests, we found that the feature space formed by such an objective function tends to grow exponentially from a certain point until it eventually explodes. We provide a more detailed analysis of this problem in Section~\ref{sec:discussion}. The solution to this problem was to find a suitable regularization that was added to the existing loss function. We decided to minimize $L_2$-norm of logits represented by functions $f$ and $g$:
\begin{equation}
\label{eq:sndstd4}
\mathcal{L}_2 = p_{\rm GL} + p_{\rm LL} = 
\sum_{h=1}^{H} \sum_{w=1}^W (\| f_{h,w} \| + \| g_{h,w} \|).
\end{equation}
Finally, we added one more regularization term $\mathcal{L}_{\sigma}$
\begin{equation}
\label{eq:sndstd5}
\mathcal{L}_{\sigma} = -\sigma(\Phi^{\rm T}(s_t))
\end{equation}
that maximizes the standard deviation $\sigma$ of the feature vector components and thus ensures that all dimensions of the feature space are used. The analysis (section~\ref{sec:analysis}) provides a more detailed justification for the introduction of this term.

The final objective function, with the scaling parameters $\beta_{1}=\beta_{2}=0.0001$ (found experimentally), was defined as
\begin{equation}
\label{eq:sndsd6}
\mathcal{L_{T}} = \frac{1}{H W} (\mathcal{L}_{\rm GL} + \mathcal{L}_{\rm LL} + \beta_{1} \mathcal{L}_{2}) + \beta_{2} \mathcal{L}_{\sigma}.
\end{equation}

\subsubsection{SND-VIC}
This method is based on the VICReg algorithm \citep{Bardes2022} belonging to the class of non-contrastive self-supervised methods. The regularization function consists of three terms: \textit{the invariant term}, which brings the feature vectors closer to each other, \textit{the variance term}, which ensures that the feature vectors within one batch have different values and \textit{the covariance term}, which ensures the decorrelation of the feature vectors and prevents information collapse. The original method does not need any negative samples, it only takes the input, creates two augmented versions, and their feature vectors are updated using the mentioned terms of the regularization function. 

Our version uses the state $s_t$ and its successor $s_{t+1}$ (the same as ST-DIM, the simple diagram can be found in Figure~\ref{fig:std_dim_idea}) instead of two augmentations of the same state.
The variance regularization term $\mathcal{L}_{v(Z)}$ is defined as a hinge function on the standard deviation of the features along the batch dimension
\begin{equation}
\label{eq:sndvic1}
\mathcal{L}_{v(Z)} = \frac{1}{D} \sum_{d=1}^{D} \max(0; \tau - \sigma(Z_d)) 
\end{equation}
where $Z_d$ is the $d$-${\rm th}$ component of all feature vectors from the batch $Z$, $\sigma$ is the actual standard deviation, and $\tau = 1$ is a constant target value for the standard deviation. This term encourages the standard deviation inside the current batch to be equal to $\tau$ along each dimension, preventing collapse with all states mapped on the same feature vector.

The covariance regularization term $\mathcal{L}_{c(Z)}$ is defined as the sum of the squared off-diagonal coefficients of the covariance matrix $C(Z)$
\begin{equation}
\label{eq:sndvic2}
\mathcal{L}_{c(Z)} = \frac{1}{(N-1)D} \sum_{i\neq j} [C(Z)]^{2}_{i,j} .
\end{equation}
This term forces feature vector components to be uncorrelated, hence enabling each feature vector component to learn unique information.

The invariance criterion $\mathcal{L}_{s(Z,Z')}$ between two batches $Z$ and $Z'$ is defined as the mean-squared Euclidean distance between each pair of feature vectors
\begin{equation}
\label{eq:sndvic3}
\mathcal{L}_{s(Z, Z')} = \frac{1}{N}\sum_{b = 1}^{N}\|Z_n - Z'_n\|^2_{2} .
\end{equation}
The overall loss $\mathcal{L}$ then takes the form
\begin{equation}
\label{eq:sndvic4}
\mathcal{L}_{T} = \lambda \mathcal{L}_{s(Z, Z')} + \mu \left[\mathcal{L}_{v(Z)} + \mathcal{L}_{v(Z')}\right] + \nu \left[\mathcal{L}_{c(Z)} + \mathcal{L}_{c(Z')}\right]
\end{equation}
where the scaling parameters are set to $\lambda=1$, $\mu=1$ and $\nu=1/25$ based on recommendations in \cite{Bardes2022}.

%%%%%%%%%%%%%%%%%%%%%%%%%%%%%%%%%%%%%%%%%%%%%%%%%%%%%%%%%%%%%%%%%%%%%%%% CHAPTER 4
\section{Experiments}
\label{sec:exper}

In this section, we focused on: 
(1) architecture search -- of the best model for distillation, answering a question of which model provides the best results; 
(2) choice of the self-supervised loss function -- which forms the distillation feature space; 
(3) hyperparameters tuning -- mostly sensitivity of choosing parameters directly related to an intrinsic motivation part.
The main experimental part is followed by an analytical part, which proves that the achieved results are a consequence of the improved internal reward function (\ref{sec:analysis_1}) and that the improvement lies in increased discriminative ability (\ref{sec:analysis_2}) and better representations (\ref{sec:analysis_3}) that use a larger part of the feature space. 
We retrained all RND baselines with our PPO model architecture (figure \ref{img:ppo_arch}), and all environment and PPO parameters remained the same (as mentioned in Tables \ref{tab:env_hyperparameters} and \ref{tab:agent_hyperparameters} ) to provide the consistent results.

Altogether we tested our methods on 10 environments (Atari and Procgen) that are considered difficult for exploration.
These include 6 Atari environments: Montezuma's Revenge, Gravitar, Venture, Private Eye, Pitfall, and Solaris. The agent receives a reward of +1 for each increase in the score, regardless of its value. It does not receive any other reward or punishment. The state (input to the models) is represented by four consecutive frames of pixels on greyscale, so the dimensionality of the state representation is 4$\times$96$\times$96. The action space is discrete, consisting of 18 actions, of which only some make sense (depending on the environment), the other actions have no impact on the environment.

We also tested 4 Procgen environments: Coinrun, Caveflyer, Jumper and Climber. Procgen is a set of procedurally generated environments, designed primarily for testing the agent's generalization \citep{cobbe2020procgen}. The paper shows several problems for generalization in RL, requiring special training and a huge number of samples. From our perspective, interesting findings (provided in Appendix B.1 of \citep{cobbe2020procgen}) are that for several seeds, the baseline agent was not able to reach a non-zero score. Those seeds lead to hard exploration environments with only a single reward at the end. Together with a fast run of these environments (thousands of FPS on a single CPU core), this makes Procgen a very good candidate for our experiments. The state is represented by two consecutive stacked frames of RGB images, so the dimensionality of the state representation is 6$\times$64$\times$64. 
The action space is discrete, consisting of 15 actions.
Preliminary experiments performed in \citep{pechac2022intrinsic} provided initial insights and led to the extensions in this work.

Before running the main experiments, we sought suitable values of hyperparameters.  For this search, we chose Montezuma's Revenge environment which represents a good compromise between game complexity and reward sparsity. This environment has been well studied (e.g. in 
\cite{aubret2019survey, burda2018exploration, bellemare13arcade, burda2018large}) and therefore provides a good comparison for tuning hyperparameter.

\subsection{Training setup}

We ran 9 simulations for each environment, taking 128M steps for Atari and 64M steps for Procgen games. Before the main training, we tried three hand-selected settings of hyperparameters for individual motivational models (mainly the scaling of the motivational signal, or regularization terms) and we always chose the one that yielded the best results. These short probes lasted 32M (Atari) or 16M (Procgen) steps and consisted of 2 or 3 simulations.

\begin{figure*}[t!]
\centering
\includegraphics[width=13cm]{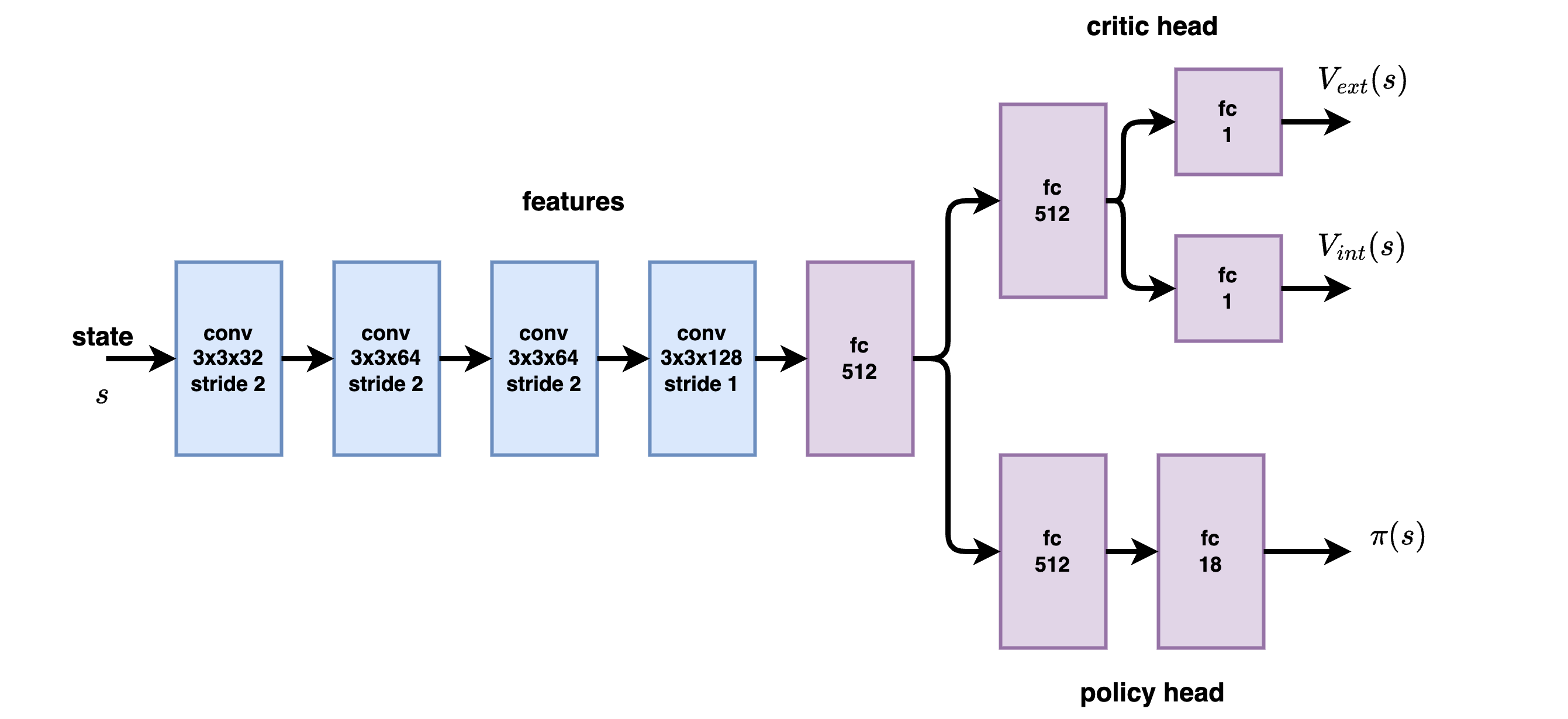}
\caption{The PPO agent model architecture.}
\label{img:ppo_arch}
\end{figure*}

\begin{figure*}[th]
\centering
\includegraphics[width=13cm]{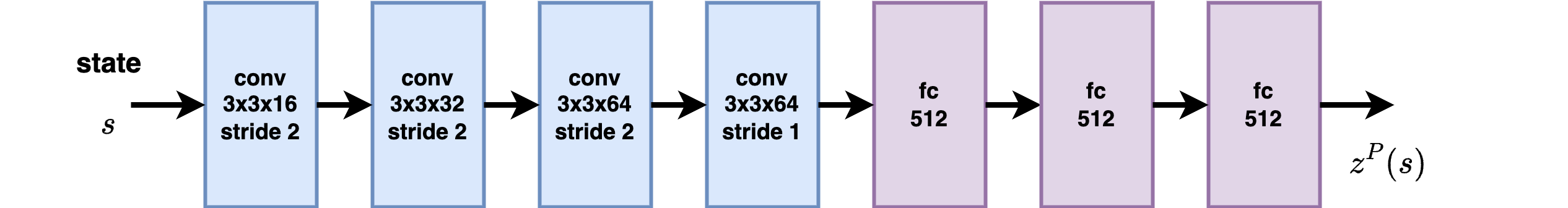}
\caption{The predictor model architecture.}
\label{img:cnd_learned_arch}
\end{figure*}

All agents were trained with the PPO algorithm \citep{schulman2017proximal} using Adam optimizer \citep{kingma2015adam} for the parameters of all modules. The basic agent consists of an actor and a critic, which are two multi-layer perceptrons sharing a common convolutional neural network (CNN) that processes the video input. The critic has two outputs (heads), one for estimating the value function for the external reward and the other for the intrinsic reward. We used the orthogonal weight initialization with a magnitude $\sqrt{2}$. 
Model architectures are presented in Figures \ref{img:ppo_arch} to %\ref{img:cnd_target_arch}, 
\ref{img:cnd_learned_arch}. 
The motivational module consists of two CNNs (the target and the learning network) which receive input from a single frame. The learning network contains two more linear layers to have an increased capacity over the target model. 

\begin{figure*}[th]
\centering
\includegraphics[width=11cm]{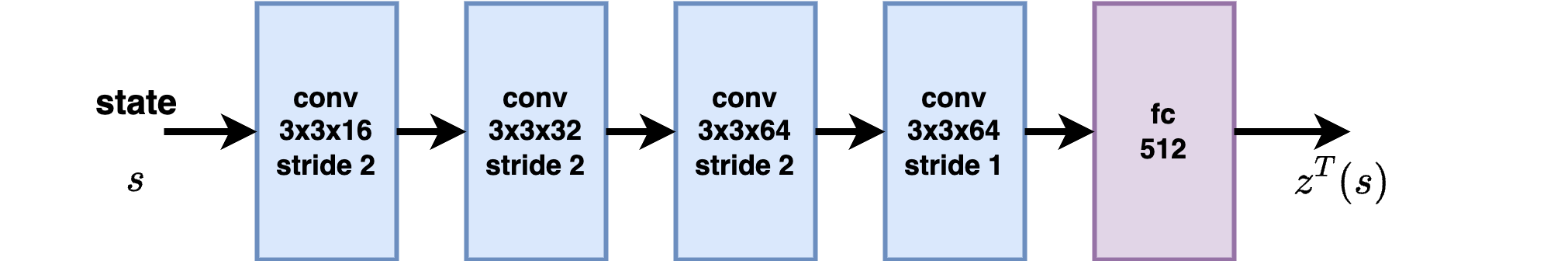}
\caption{The target model architecture.}
\label{img:cnd_target_arch}
\end{figure*}

\begin{table}[th]
\scriptsize
\centering
\caption{Environment hyperparameters}
  \begin{tabular}{l|ll}
     Hyperparameter & Atari & Procgen  \\ 
      \hline\hline
      Observation downsampling & 96$\times$96 & 64$\times$64 \\
      Frame stacking & 4 & 2 \\
      State dimensionality for PPO & 4$\times$96$\times$96 & 6$\times$64$\times$64 \\
      State dimensionality for IM modules & 1$\times$96$\times$96 & 3$\times$64$\times$64 \\
      Parallel environments count & 128 & 128 \\
      State normalisation & $s/255$ & $s/255$ \\
      Samples per environment & 1M & 0.5M \\
      \hline
  \end{tabular}\label{tab:env_hyperparameters}
\end{table}

\begin{table}[thb]
\scriptsize
\centering
\caption{Agent hyperparameters}
    \begin{tabular}{l|l}
        Hyperparameter & Value \\ 
        \hline\hline
        PPO model learning rate & $0.0001$ \\
        Target model $\Phi^{\rm T}$ learning rate & $0.0001$ \\
        Predictor model $\Phi^{\rm P}$ learning rate & $0.0001$ \\
        Discount factor $\gamma^{\rm ext}$ & $0.998$ \\
        Discount factor $\gamma^{\rm intr}$ & $0.99$ \\
        Advantage external coefficient & $2.0$ \\
        Advantage intrinsic coefficient & $1.0$ \\
        Intrinsic reward scaling & $0.5$ \\
        Rollout length & $128$ \\ 
        Number of optimization epochs & $4$ \\
        Entropy coefficient & $0.001$ \\
        Epsilon clipping & $0.1$ \\
        Gradient norm clipping & $0.5$ \\
        GAE $\lambda$ coefficient & $0.95$ \\
        Optimizer & Adam \\
        Weight initialisation & orthogonal, gain = $\sqrt{2}$ \\
        \hline
    \end{tabular}
    \label{tab:agent_hyperparameters}
\end{table}

\begin{table}[th]
\scriptsize
\centering
\caption{Average cumulative reward (with standard deviation) per episode for all 3 preprocessing methods and maximal reward achieved by the agents, in case of Montezuma's Revenge task.}
\begin{tabular}{l|ll}
Preprocessing method & Average reward & Max. reward \\
\hline\hline
normalization    & 3.60 $\pm$ 0.14 & 7 \\ 
mean subtraction & 4.13 $\pm$ 0.12 & 7 \\
none & 2.31 $\pm$ 0.20 & 7 \\
\hline
\end{tabular}
\label{tab:res1}
\end{table}

We followed \citep{burda2018exploration} for setting the hyperparameters, to allow consistent comparison of results. 
We ran 128 parallel environments.
For Atari, we used 1M samples for each environment  (total 128M frames), for Procgen we used 0.5M samples for each environment (total 64M frames). 
The intrinsic motivation modules used no frame stacking, a single grayscale image for Atari environments and a single RGB image for Procgen.

The summary of all environment hyperparameters is in Table \ref{tab:env_hyperparameters}.
The discount factors were set to $\gamma^{\rm ext} = 0.998$ for external reward and $\gamma^{\rm intr} = 0.99$ for intrinsic reward.
The simulations revealed the importance of intrinsic reward scaling, the best result was achieved for $\eta = 0.5$.
The learning rate for all models was set to 0.0001. Actor and critic models used ReLU and motivation models worked best with ELU activation function. We also found that the deeper model with 3$\times$3 convolutions worked better than the standard Atari model using 8$\times$8 or 4$\times$4 convolutions in \citep{mnih2013playing}. We retrained RND models to obtain comparable results and found faster convergence. The summary of PPO agent hyperparameters is in Table \ref{tab:agent_hyperparameters}.
More hyperparameters and further details of the learning process and the architectures of modules can be found in our source codes.

\subsection{State preprocessing}
\label{sec:exp2}

The state before entering the motivation module of the SND model can undergo preprocessing. We tested three preprocessing methods:
\begin{enumerate}
\item State normalization using the running mean and standard deviation, 
\item Subtraction of the running mean value from the state, 
\item No preprocessing.
\end{enumerate}

We performed two training runs for each preprocessing method in 32M steps. For testing, we used the SND-STD method. Table~\ref{tab:res1} demonstrates that the state preprocessing did not have a significant effect on agent performance (maximum reward achieved), only on the speed of learning. This also agrees with our assumption that operations such as subtraction of the mean or normalization should be able to find the network itself, trained using the self-supervised loss function. Therefore, it is not necessary for the designer to put them into the learning process explicitly. These conclusions will still need to be confirmed by statistical analysis. RND used mean subtraction and SND-V, SND-STD together with SND-VIC did not use input state preprocessing.

\begin{figure*}[t!]
\centering
\includegraphics[width=10cm]{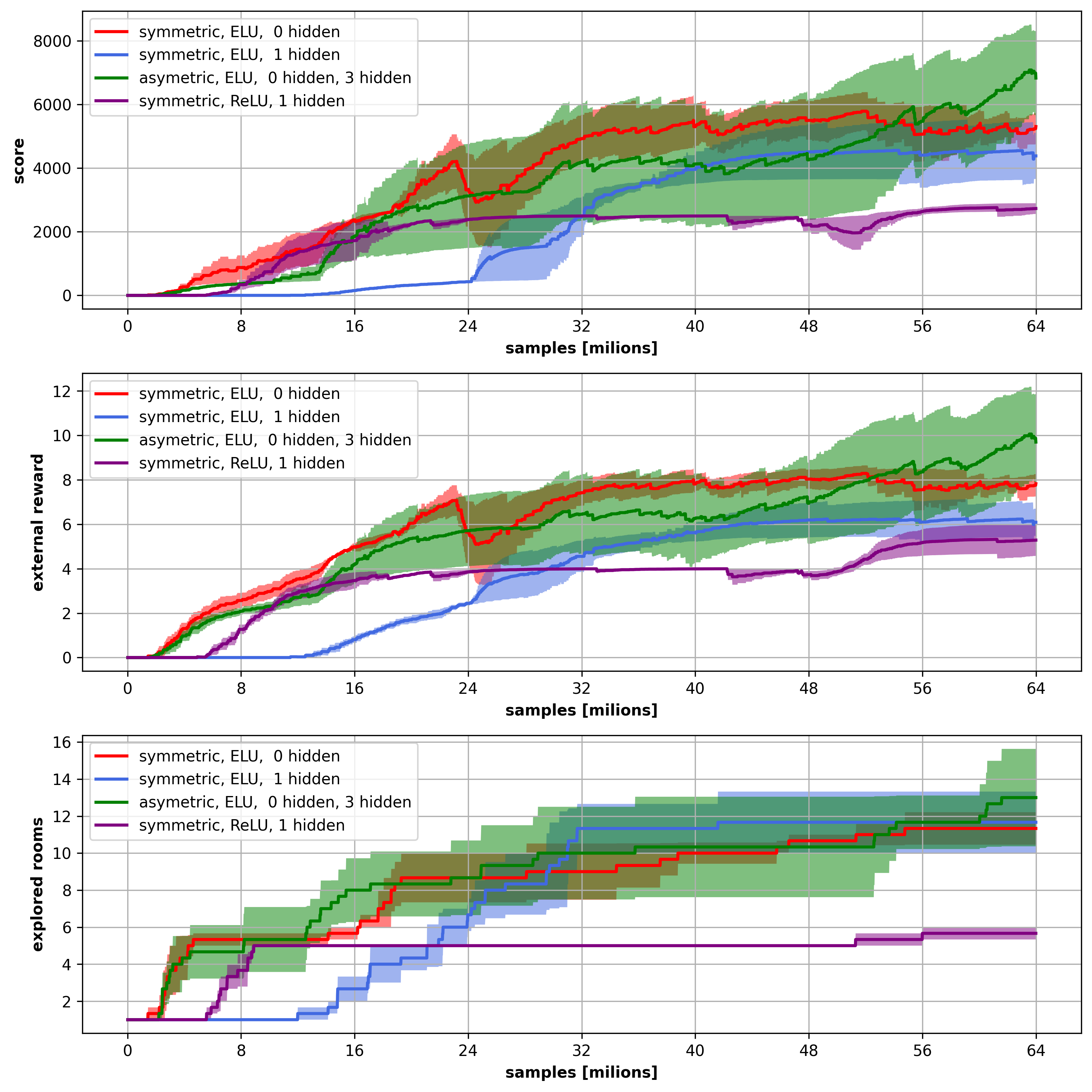}
\caption{Agent performance based on various predictor model architectures, in case of Montezuma’s Revenge task, evaluated in terms of the overall score, external reward obtained and the number of rooms explored.}
\label{img:result_cnd_learned_arch}
\end{figure*}

\subsection{Results}

We processed several quick experiments on Montezuma's Revenge to explore an optimal setup.
First, we have to test the optimal architecture of $\Phi^{\rm T}$ and $\Phi^{\rm P}$ models. 
We experimented with 4 architectures: 
\begin{enumerate}
    \item identical models, one fully connected output layer, ELU activations
    \item identical models, two fully connected layers, ELU activations
    \item identical models, one fully connected layer, ReLU activations
    \item asymmetric models, three fully connected layers for $\Phi^{\rm P}$, none for $\Phi^{\rm T}$, ELU activations fixed to fully connected convention.
\end{enumerate}
Results of different model architectures are in Figure~\ref{img:result_cnd_learned_arch}. The best result was achieved for the asymmetric architecture.

Next, we tested the effect of different augmentations. We considered three scenarios: 
\begin{enumerate}
    \item uniform noise, $\langle -0.2, 0.2 \rangle$
    \item uniform noise, random tiles masking (tiles with sizes 1, 2, 4, 8, 12, 16)
    \item uniform noise,  random tiles masking, random convolution filter apply.
\end{enumerate}

\begin{figure*}[t!]
\centering
\includegraphics[width=10cm]{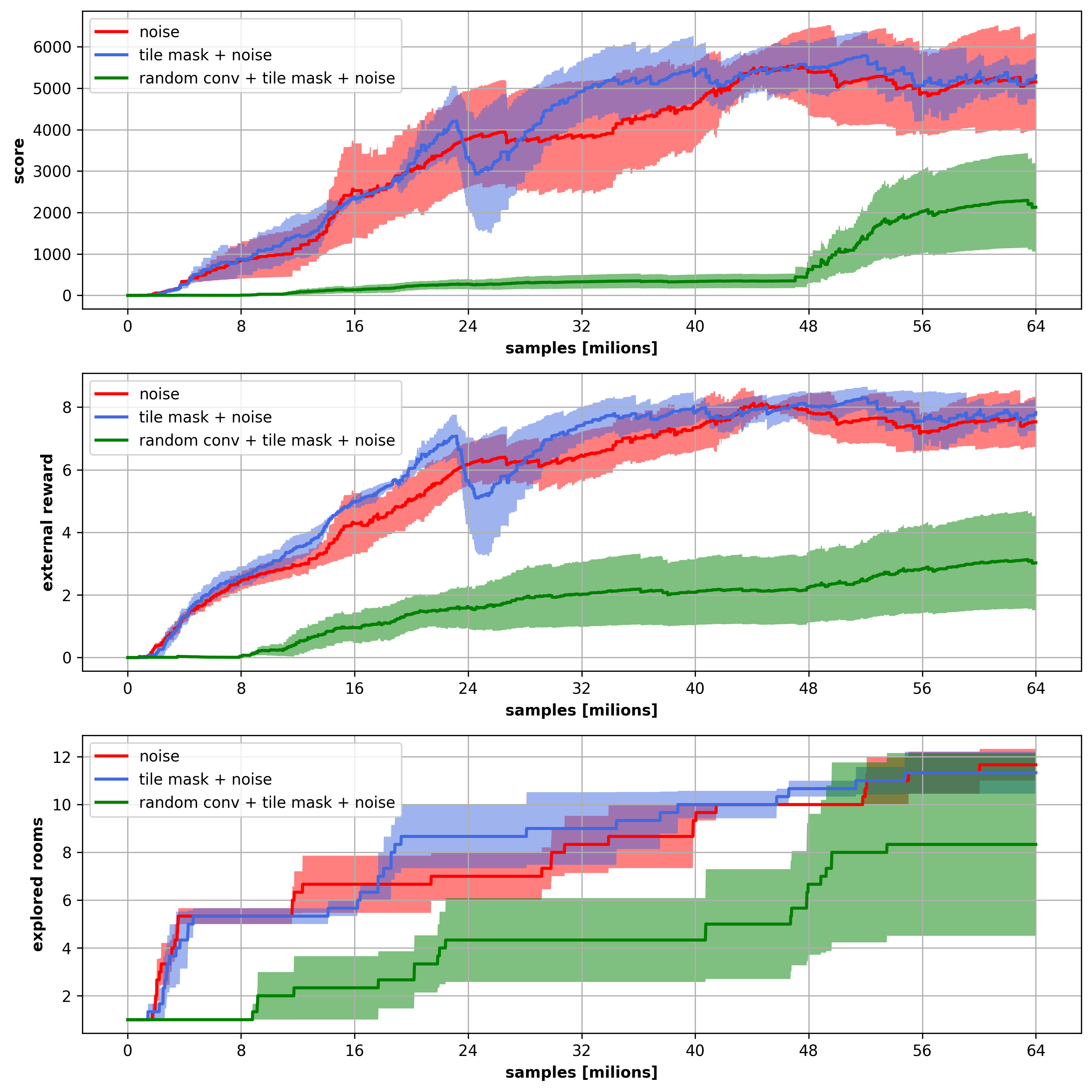}
\caption{Agent performance for different state augmentations, in case of Montezuma’s Revenge task, evaluated in terms of the overall score, external reward obtained and the number of rooms explored.}
\label{img:result_cnd_aug}
\end{figure*}

Noise augmentation is commonly used in supervised image training. The tile masking forces the model to reconstruct incomplete information. Random convolution filter
helps the model to learn to focus on informative features, not on texture colors. The results of different state augmentations are in Figure~\ref{img:result_cnd_aug}.
We hypothesized that all three augmentations together would provide the best results. However, probably due to a shallow target model, this was not the case since the model was not able to learn sufficient transformation invariants. The best result was using uniform noise alone or with random tile masking as augmentations, in the case of shallow target models.

\begin{figure*}[t!]
\centering
\includegraphics[width=10cm]{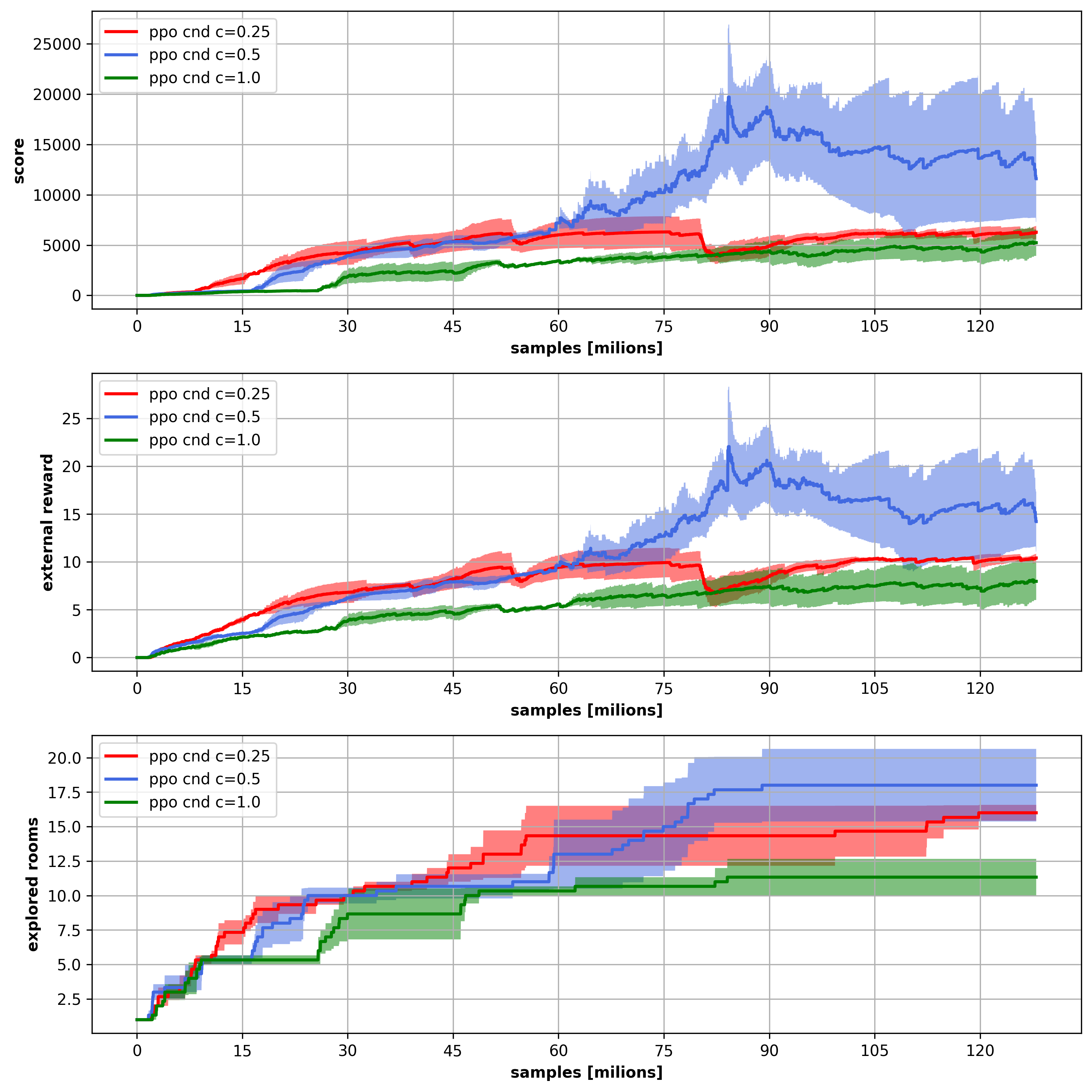}
\caption{Agent performance for different intrinsic reward scaling methods, in case of Montezuma’s Revenge task, evaluated in terms of the overall score, external reward obtained and the number of rooms explored.}
\label{img:result_cnd_scaling}
\end{figure*}

\begin{figure}[t!]
  \centering
  \begin{subfigure}[b]{0.32\textwidth}
    \centering
    \includegraphics[width=4.1cm]{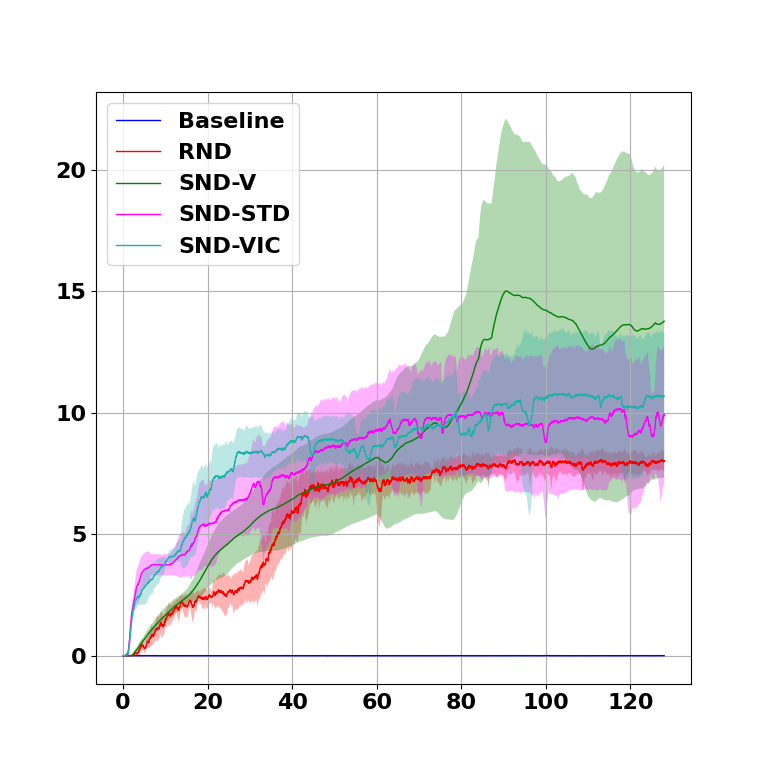}
    \caption{Montezuma's Revenge}
    \label{fig:res2a}
  \end{subfigure}
  \begin{subfigure}[b]{0.32\textwidth}
    \centering
    \includegraphics[width=4.1cm]{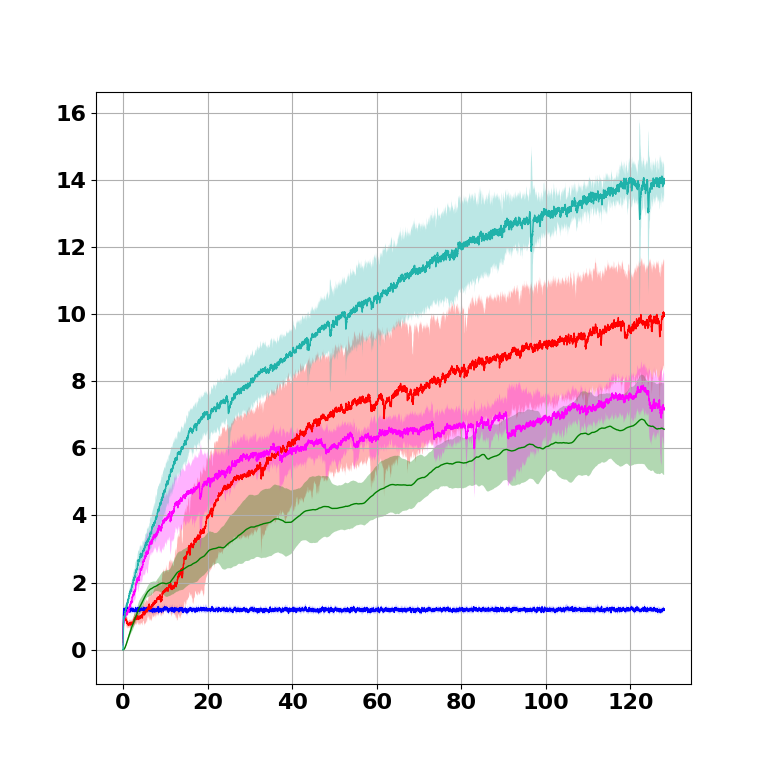}
    \caption{Gravitar}
    \label{fig:res2b}
  \end{subfigure}  
  \begin{subfigure}[b]{0.32\textwidth}
    \centering
    \includegraphics[width=4.1cm]{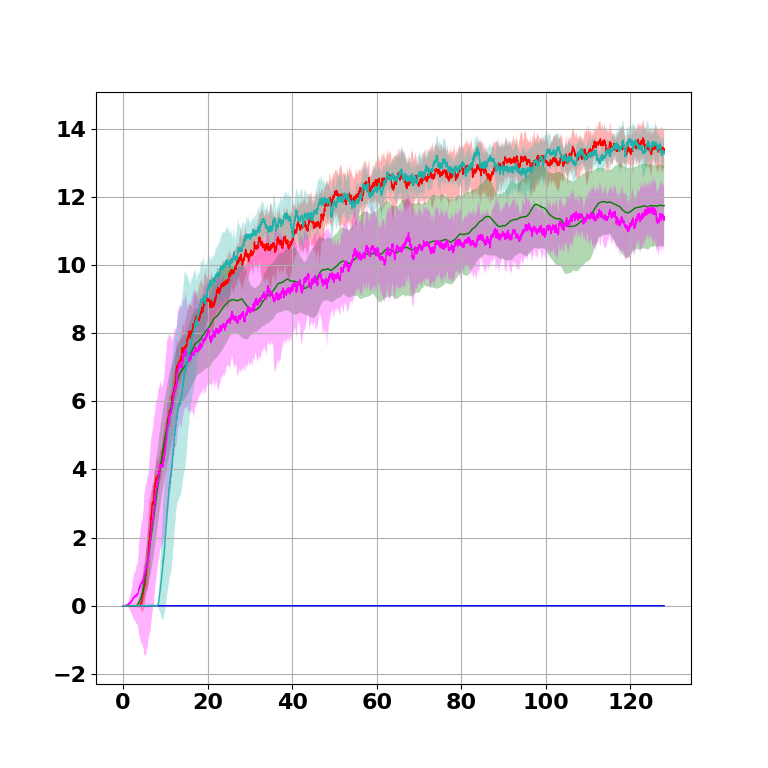}
    \caption{Venture}
    \label{fig:res2c}
  \end{subfigure}
  \begin{subfigure}[b]{0.32\textwidth}
    \centering
    \includegraphics[width=4.1cm]{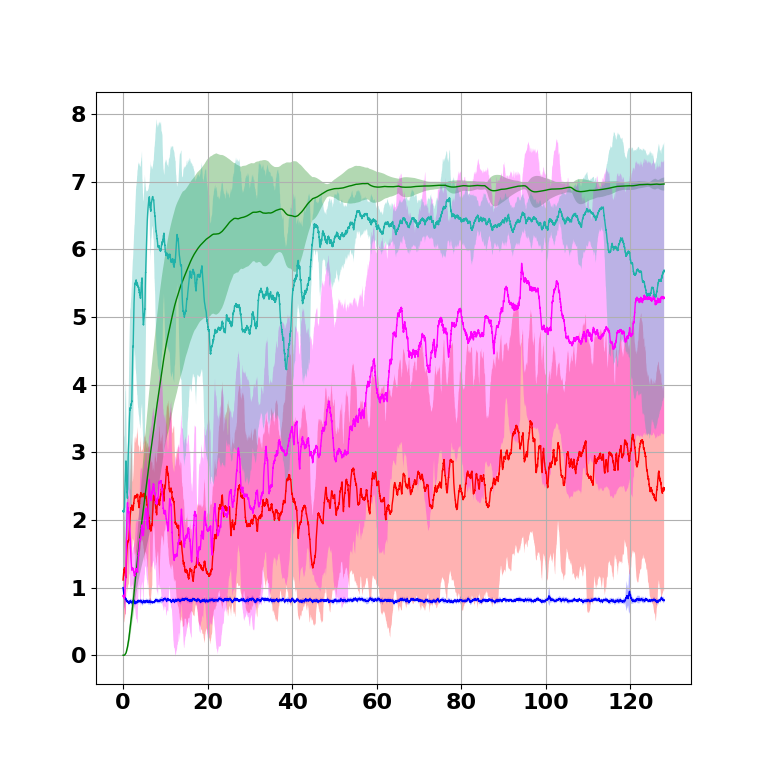}
    \caption{Private Eye}
    \label{fig:res2d}
  \end{subfigure} 
  \begin{subfigure}[b]{0.32\textwidth}
    \centering
    \includegraphics[width=4.1cm]{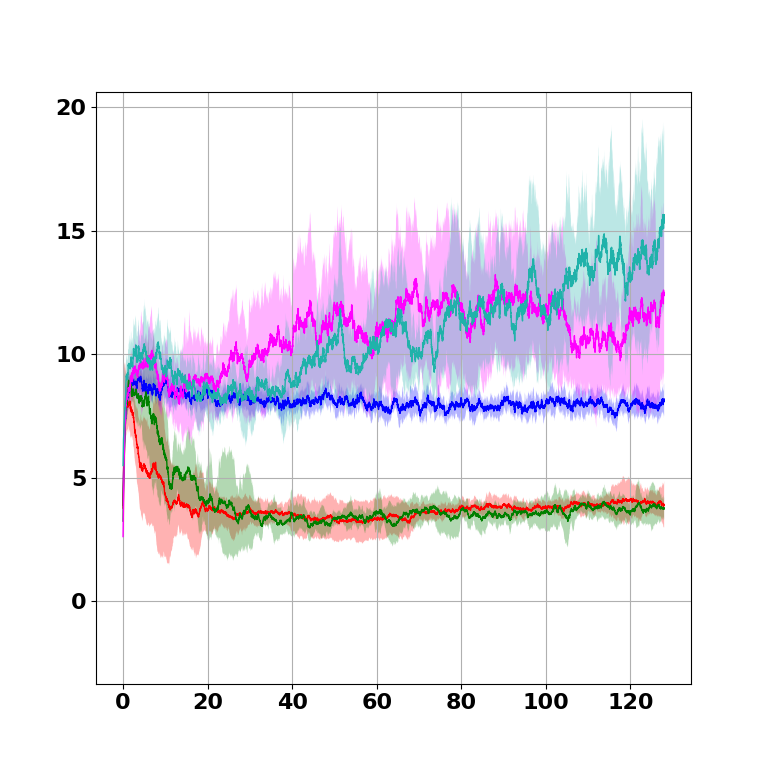}
    \caption{Solaris}
    \label{fig:res2f}
  \end{subfigure}
    \begin{subfigure}[b]{0.32\textwidth}
    \centering
    \includegraphics[width=4.1cm]{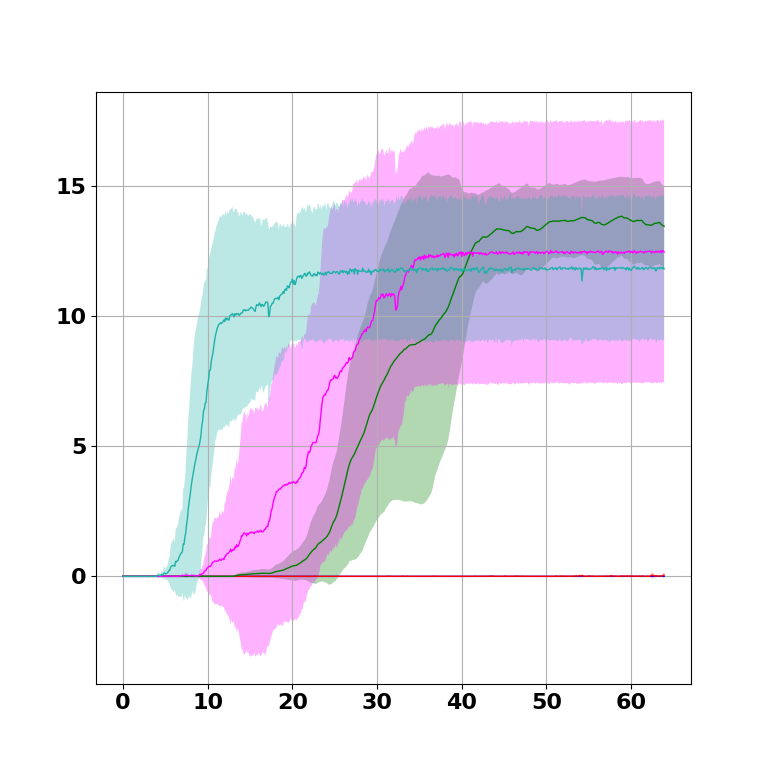}
    \caption{Caveflyer}
    \label{fig:res2g}
  \end{subfigure}
    \begin{subfigure}[b]{0.32\textwidth}
    \centering
    \includegraphics[width=4.1cm]{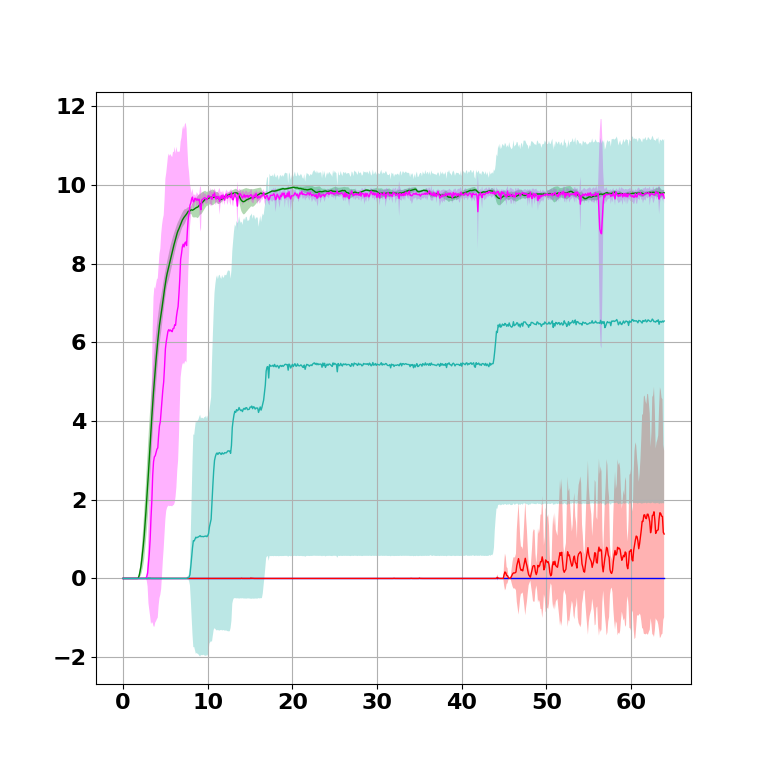}
    \caption{Coinrun}
    \label{fig:res2h}
  \end{subfigure}
    \begin{subfigure}[b]{0.32\textwidth}
    \centering
    \includegraphics[width=4.1cm]{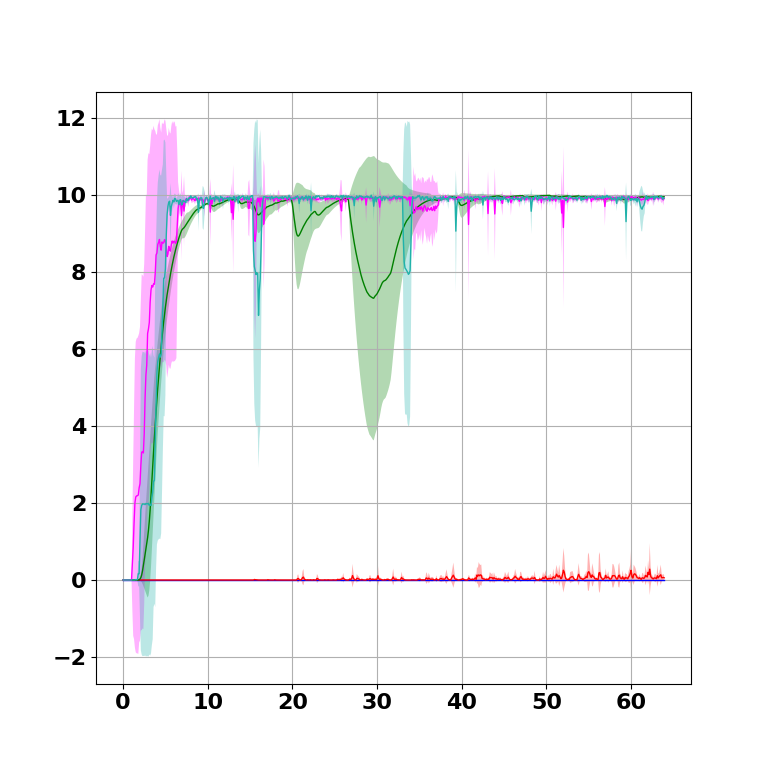}
    \caption{Jumper}
    \label{fig:res2i}
  \end{subfigure}
    \begin{subfigure}[b]{0.32\textwidth}
    \centering    
    \includegraphics[width=4.1cm]{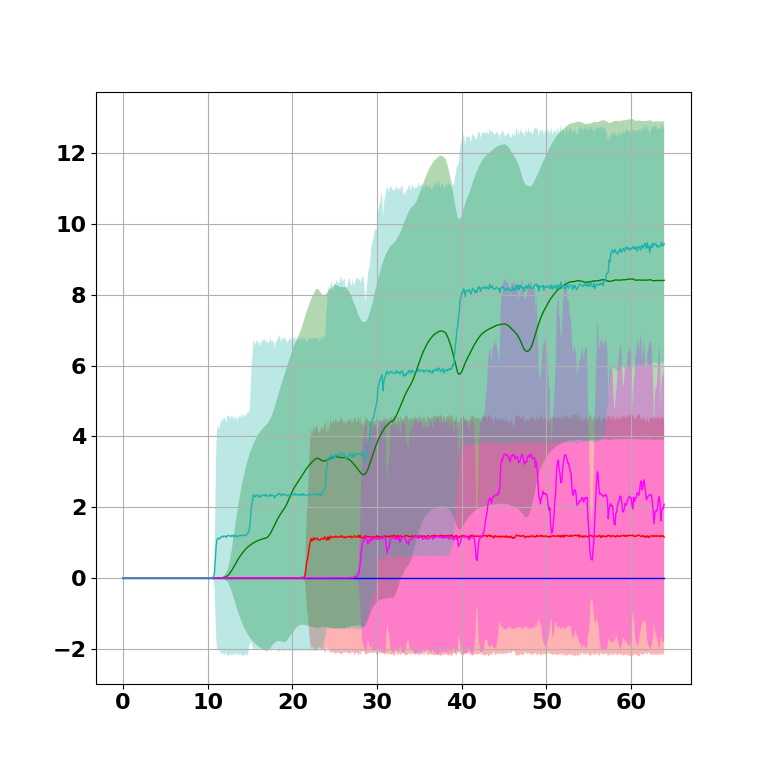}
\caption{Climber} \label{fig:res2j}
  \end{subfigure}
\caption{The cumulative external reward per episode (with the standard deviation) received by the agent from the tested environment. We omitted the graph for the Pitfall environment, where all algorithms were not successful, having achieved zero reward. The horizontal axis shows the number of steps in millions, and the vertical axis refers to the external reward.}
\label{fig:result}
\end{figure}

\begin{table}[thb]
\scriptsize
\centering
\caption{Average cumulative external reward per episode for tested models. The best models for each environment are shown in boldface.}
\begin{tabular}{l|cccccc}
\hline
 & Baseline & RND & SND-V & SND-STD & SND-VIC \\
\hline\hline
Montezuma & 0.00 $\pm$ 0.00 & 5.33 $\pm$ 0.23 & \textbf{8.87 $\pm$ 2.79} & \textbf{7.76 $\pm$ 1.73} & \textbf{8.45 $\pm$ 1.12}  \\
Gravitar & 1.19 $\pm$ 0.00 & 6.63 $\pm$ 1.55 & 4.73 $\pm$ 0.61 & 5.89 $\pm$ 0.43 & \textbf{10.05 $\pm$ 0.66} \\
Venture & 0.00 $\pm$ 0.00 & \textbf{11.18 $\pm$ 0.42} & 9.71 $\pm$ 0.97 & 9.54 $\pm$ 0.90 & \textbf{11.36 $\pm$ 0.37} \\ 
Private Eye & 0.81 $\pm$ 0.01 & 2.41 $\pm$ 0.95 & \textbf{6.44 $\pm$ 0.27} & 3.79 $\pm$ 1.24 & 5.93 $\pm$ 0.47 \\ 
Pitfall & 0.00 $\pm$ 0.00 & 0.00 $\pm$ 0.00 & 0.00 $\pm$ 0.00 & 0.00 $\pm$ 0.00 & 0.00 $\pm$ 0.00 \\ 
Solaris & 8.08 $\pm$ 0.15 & 3.84 $\pm$ 0.25 & 3.96 $\pm$ 0.41 & \textbf{11.61 $\pm$ 1.12} & \textbf{10.85 $\pm$ 1.20}  \\
Caveflyer & 0.00 $\pm$ 0.00 & 0.00 $\pm$ 0.00 & 7.28 $\pm$ 1.62 & \textbf{10.86 $\pm$ 4.37} & \textbf{11.14 $\pm$ 2.35} \\
Coinrun & 0.00 $\pm$ 0.00 & 0.25 $\pm$ 0.50 & \textbf{9.40 $\pm$ 0.05} & \textbf{9.40 $\pm$ 0.07} & 2.55 $\pm$ 3.63  \\
Jumper & 0.00 $\pm$ 0.00 & 0.03 $\pm$ 0.02 & 9.22 $\pm$ 0.21 & \textbf{9.76 $\pm$ 0.04} & \textbf{9.76 $\pm$ 0.03}\\
Climber & 0.00 $\pm$ 0.00 & 1.00 $\pm$ 2.83 & 4.63 $\pm$ 2.96 & 1.48 $\pm$ 2.40 & \textbf{6.62 $\pm$ 3.24} \\
\hline
\end{tabular}
\label{tab:res2}
\end{table}

\begin{table}[htb]
\scriptsize
\centering
\caption{Results of two-sided nonparametric Mann-Whitney U tests on selected pairs of models.}
\begin{tabular}{l|cccc}
\hline
 Environment & Method & Method & U statistic & $p$-value \\
\hline\hline
Montezuma & SND-V & SND-STD & 45.00 & 0.42 \\
& SND-V & SND-VIC & 38.00 & 0.89 \\
& SND-STD & SND-VIC & 25.00 & 0.51 \\
Venture & RND & SND-VIC & 27.00 & 0.42 \\
Private Eye & SND-V & SND-VIC & 63.00 & 0.01 \\
Solaris & SND-STD & SND-VIC & 34.00 & 0.88 \\
Caveflyer & SND-STD & SND-VIC & 40.00 & 1.00 \\
Coinrun & SND-V & SND-STD & 14.00 & 1.00 \\
Jumper & SND-V & SND-STD & 0.00 & 0.01 \\
& SND-V & SND-VIC & 0.00 & 0.04 \\
& SND-STD & SND-VIC & 25.00 & 0.80 \\
\hline
\end{tabular}
\label{tab:mannwhitneyu}
\end{table}

\begin{table}[htb]
\scriptsize
\centering
\caption{Average maximal score reached by tested models on Atari environments. The best model for each environment is shown in boldface.}
\begin{tabular}{l|cccccc}
\hline
 & Baseline & RND & SND-V & SND-STD & SND-VIC \\
\hline\hline
Montezuma & 400 & 6689 & \textbf{14973} & 7212 & 7838 \\
Gravitar & 2611 & 5600 & 2469 & 4643 & \textbf{6712} \\
Venture & 22 & 2167 & 1661 & 2138 & \textbf{2188}  \\ 
Private Eye & 14870 & 14996 & 4438 & 15089 & \textbf{17313} \\ 
Pitfall & 0 & 0 & 0 & 0 & 0 \\ 
Solaris & 12344 & 10667 & 11582 & \textbf{12460} & 11865 \\ 
\hline
\end{tabular}
\label{tab:res3}
\end{table}

Finally, we tested an intrinsic reward scaling. A low value can lead to stacking the agent into a non-exploring policy. High value can prevent the agent from collecting extrinsic rewards, or make it too sensitive to small unimportant changes, both causing instability. We tested three values: 0.25, 0.5 and 1.0. Figure~\ref{img:result_cnd_scaling} shows that the best score is achieved with a 0.5 reward scaling value. However, some environments provide better results after fine-tuning to 0.25.
Figure~\ref{fig:result} captures the cumulative external reward per episode and the standard deviation of the tested models in 9 different environments. 

Table~\ref{tab:res2} provides the overall average results on tested models in 10 environments.
It is evident from visual inspection that our SND models excel in the case of all but one environment. For each environment, the best models (with the highest average score) are shown in bold to facilitate orientation (as explained below). In order to get a statistically more accurate view of the results\footnote{We agree with the recently proposed view \citep{Agarwal2021} that it is important to go beyond comparing point estimates of aggregate performance (while neglecting statistical uncertainty) when evaluating and comparing the performance of RL models.}, we applied statistical tests that we found meaningful in the case of our data. 
Based on a smaller number of (time intensive) runs (8-9) for each environment and the model, and large differences between models (in most cases) we applied paired non-parametric Mann-Whitney U tests for models with similar (top) performance to test significance of differences. The results shown in Table~\ref{tab:mannwhitneyu}, provide a $p$-value for each paired test. Higher $p$-values indicate that in most environments (Montezuma, Venture, Solaris, Caveflyer, Coinrun, Jumper) the top-performing models were statistically indifferent (and hence are all shown in bold in Table~\ref{tab:res2}).

Table~\ref{tab:res3} shows the maximum achieved score for Atari environments\footnote{Providing the maximum scores has been the historical tradition in the case of Atari games, probably motivated by an expected increase of their attractiveness.}
which is often used for model comparison (although we would like to emphasize that the agent never receives this score as a reward, and therefore it is not its goal to maximize it).
%\sout{Among the tested environments, Pitfall task exceeded the capabilities of all tested algorithms, since none of them achieved a single reward point. In the remaining 9 environments,} 
The best results were achieved with the models based on SND motivation, while in 8 cases it was with a significant lead over the existing algorithms (in the Venture environment the results were almost the same as those of the RND model). When evaluating the score, the SND models achieved the highest score in 5 Atari environments (except the already mentioned Pitfall), and in 3 cases (Montezuma's Revenge, Gravitar, Private Eye) it was significantly higher than the compared models.

In a Venture environment, the RND model scores comparably to the best SND-VIC model. The reason could be that all SND  agents fall into a ``repetitive reward trap'': the agent quickly learns to enter one room, collect points, and after exiting, enter again where points can be collected again by the same policy. 
The Pitfall environment looks most challenging, due to extremely sparse rewards, which probably led to the complete failure of all models. It contains very similar rooms, where the noise dominates - as NPC creatures blink or randomly move, which misleads the agents into focusing on noise as a novelty.

\begin{figure*}[t!]
  \begin{subfigure}{1.0\textwidth}
    \centering\includegraphics[width=12cm]{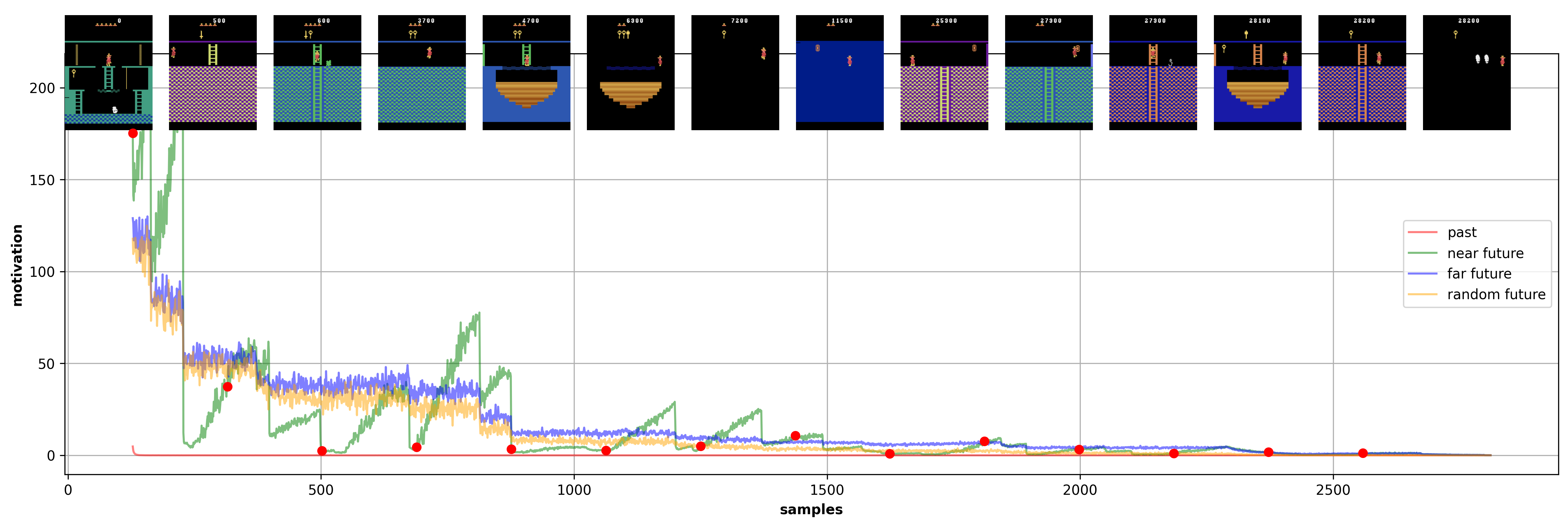}
    \caption{RND}
    \label{fig:nov_rnd_result_summary}
   \end{subfigure}
  \\
  \begin{subfigure}{1.0\textwidth}
    \centering\includegraphics[width=12cm]{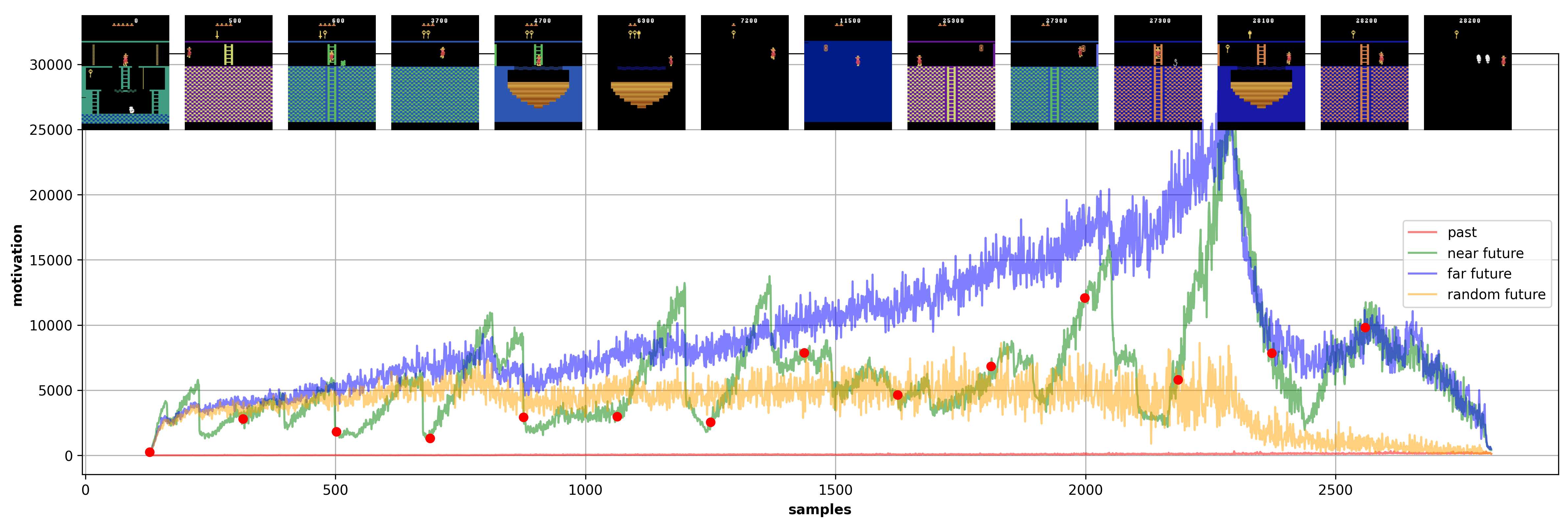}
    \caption{SND-STD}
    \label{fig:nov_nce_result_summary}
   \end{subfigure}
  \\
  \begin{subfigure}{1.0\textwidth}
    \centering\includegraphics[width=12cm]{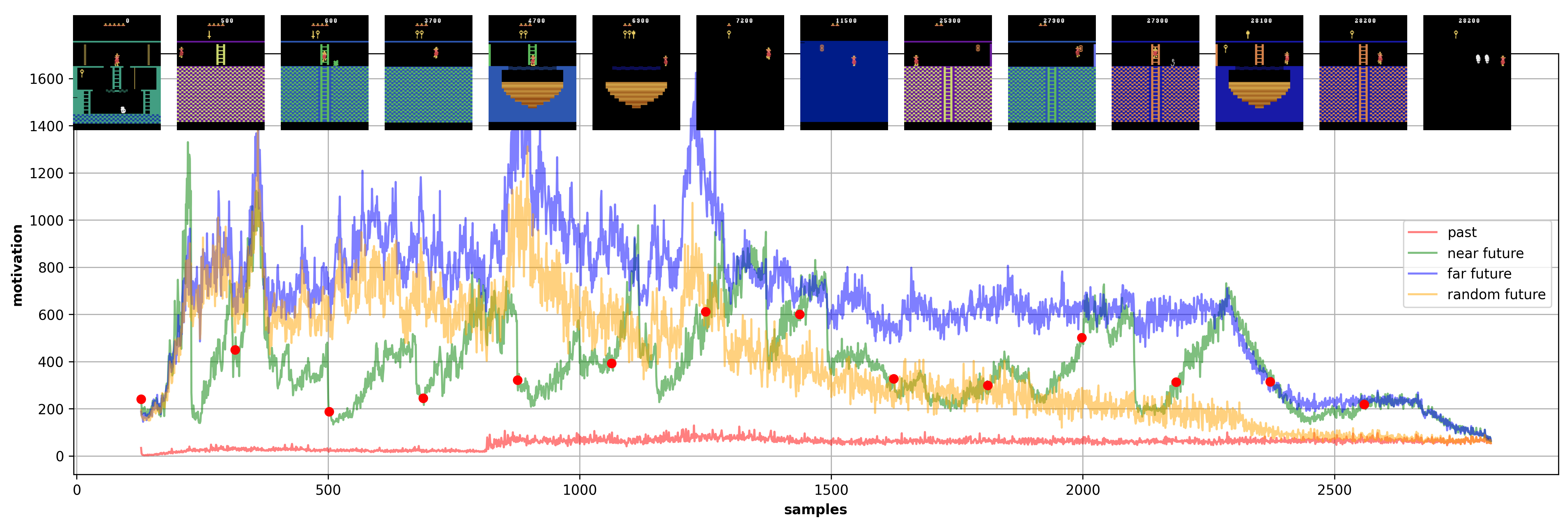}
    \caption{SND-V}
    \label{fig:nov_mse_result_summary}
   \end{subfigure}
  \\
  \begin{subfigure}{1.0\textwidth}
    \centering\includegraphics[width=12cm]{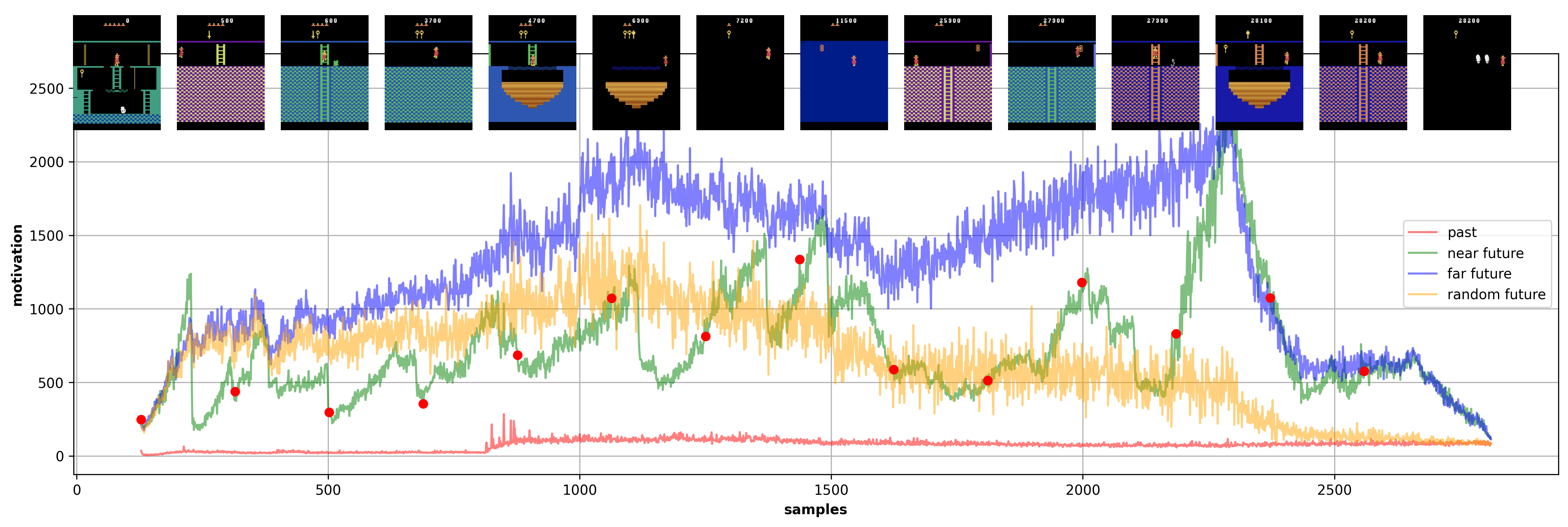}
    \caption{SND-VIC}
    \label{fig:nov_vicreg_result_summary}
   \end{subfigure}
\caption{Novelty detection for different regularisation losses as a reaction to different future windows. The states were collected on Montezuma's Revenge with our best agent, red dots correspond to state examples shown above.}

\end{figure*}

\begin{figure}[t!]
  \centering
  \begin{subfigure}[b]{0.35\textwidth}
    \centering
    \includegraphics[width=4.4cm]{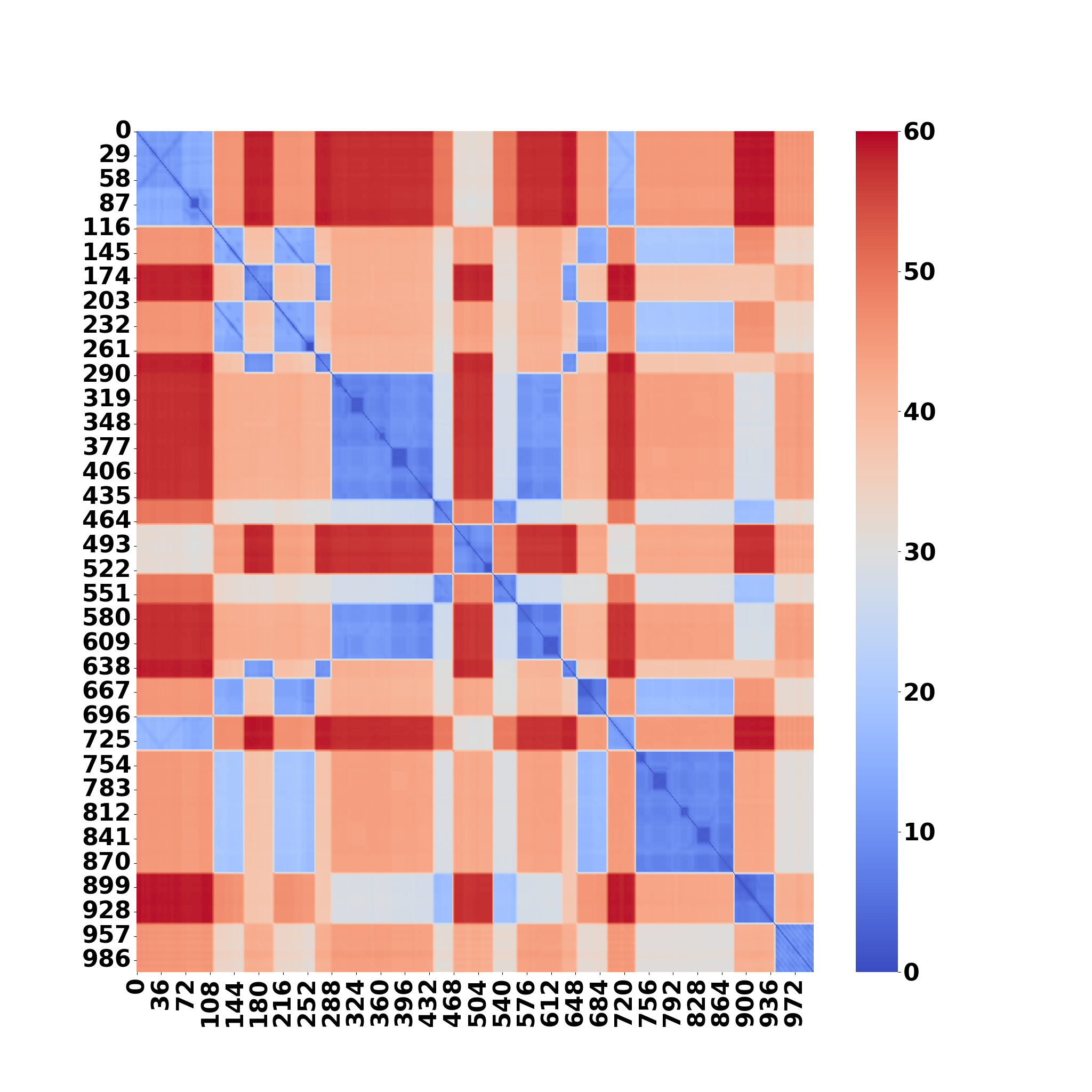}
    \caption{States}
    \label{fig:analysis2a}
  \end{subfigure}
  \begin{subfigure}[b]{0.35\textwidth}
    \centering
    \includegraphics[width=4.4cm]{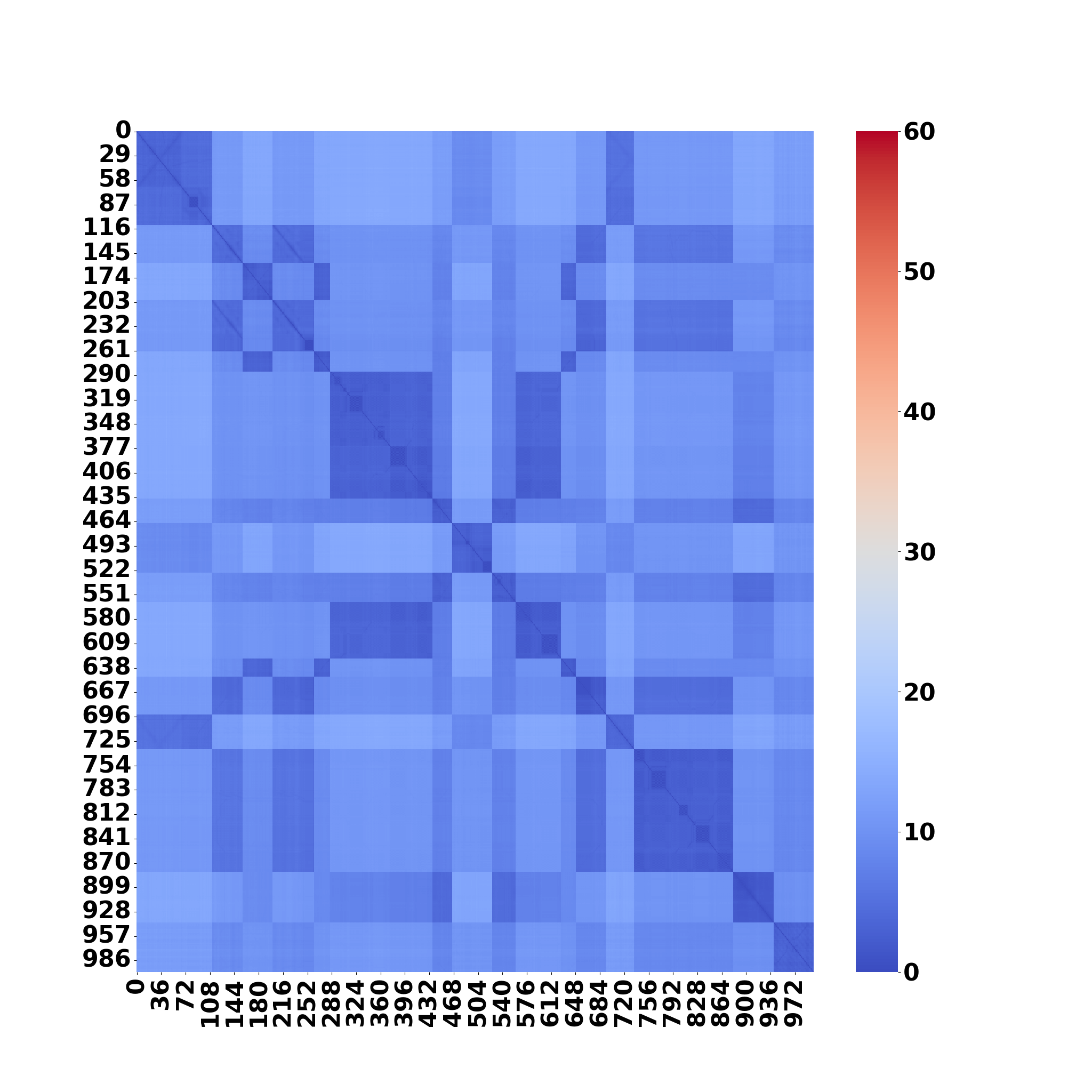}
    \caption{RND}
    \label{fig:analysis2b}
  \end{subfigure}  
  \begin{subfigure}[b]{0.3\textwidth}
    \centering
    \includegraphics[width=4.4cm]{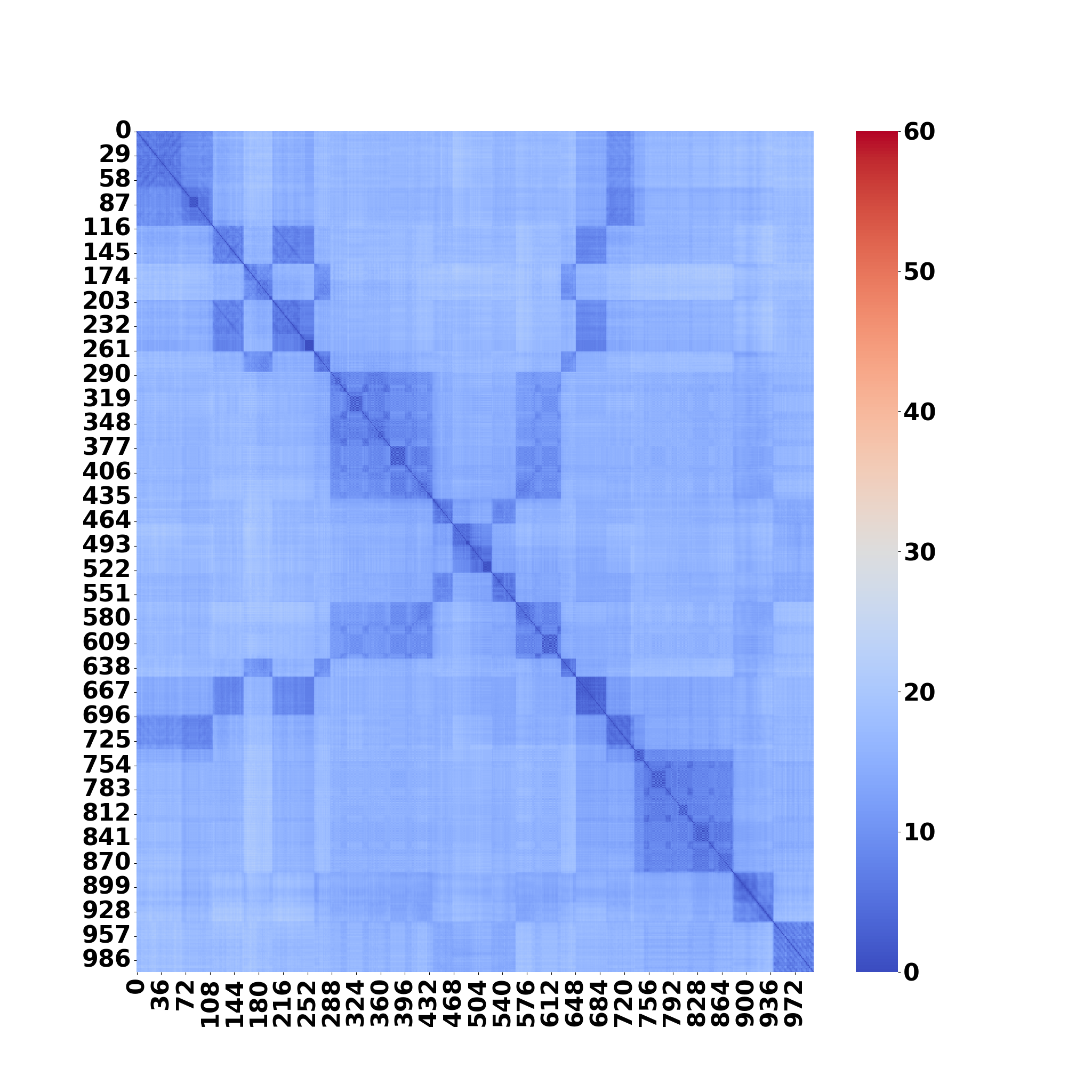}
    \caption{SND-STD}
    \label{fig:analysis2d}
  \end{subfigure}
  \begin{subfigure}[b]{0.3\textwidth}
    \centering
    \includegraphics[width=4.4cm]{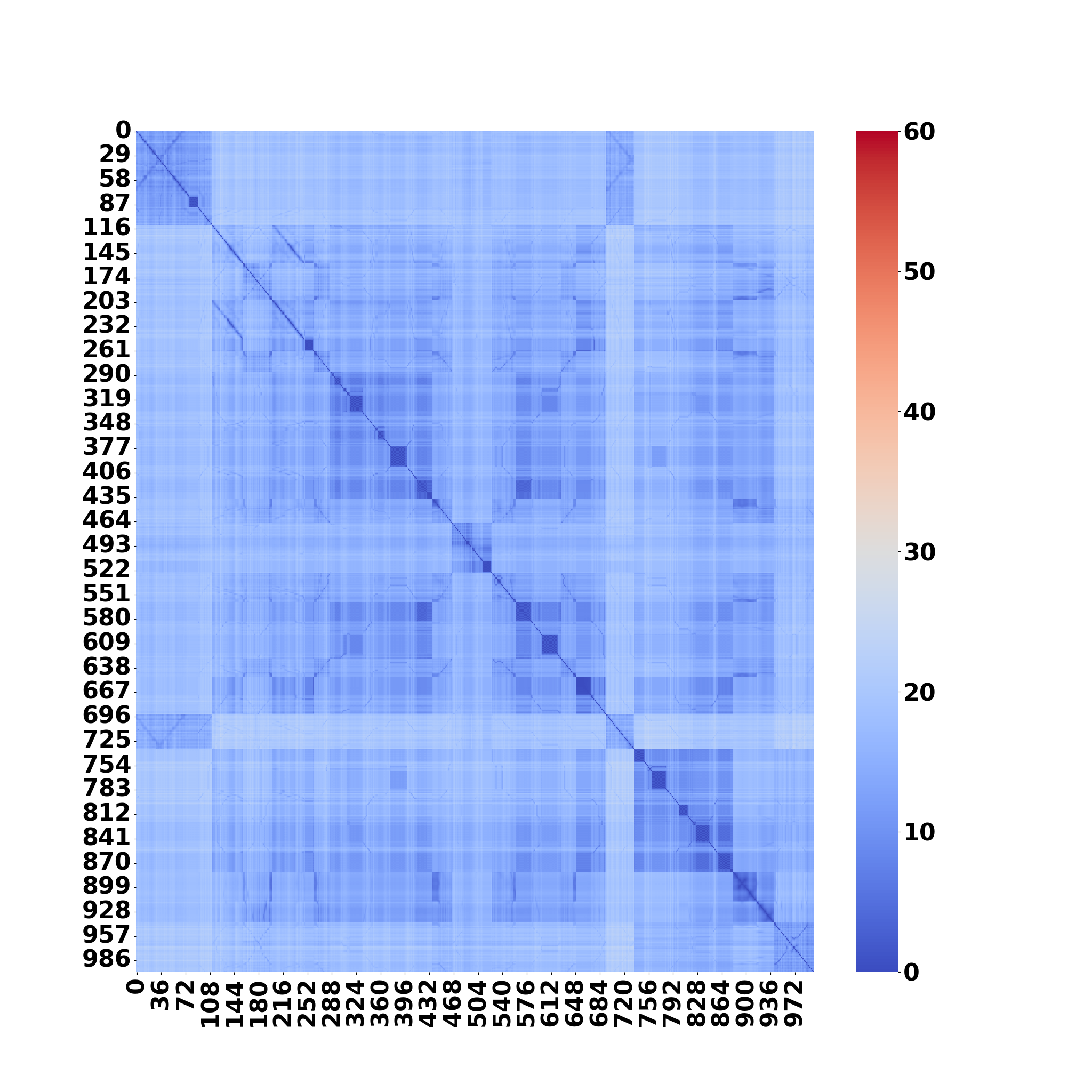}
    \caption{SND-V}
    \label{fig:analysis2c}
  \end{subfigure}
  \begin{subfigure}[b]{0.3\textwidth}
    \centering
    \includegraphics[width=4.4cm]{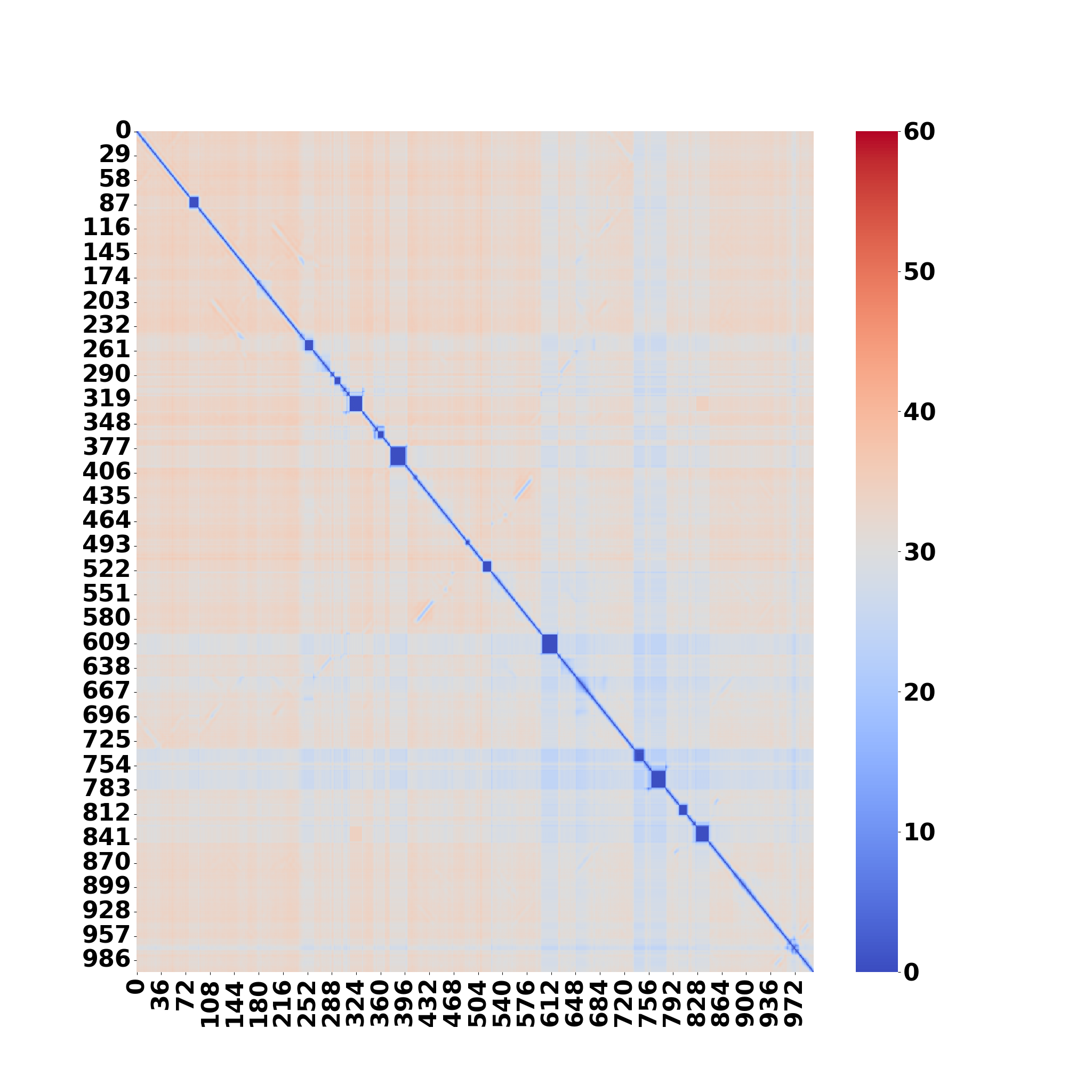}
    \caption{SND-VIC}
    \label{fig:analysis2e}   
  \end{subfigure}
\caption{The distance matrix of 1000 states and the feature vectors collected from Montezuma's Revenge environment. Small distances are displayed in blue color, while large distances are displayed in red color.}
\label{fig:snd_state_features}
\end{figure}

\begin{figure}[thb]
  \begin{subfigure}{0.5\textwidth}
    \centering
    \includegraphics[width=7cm]{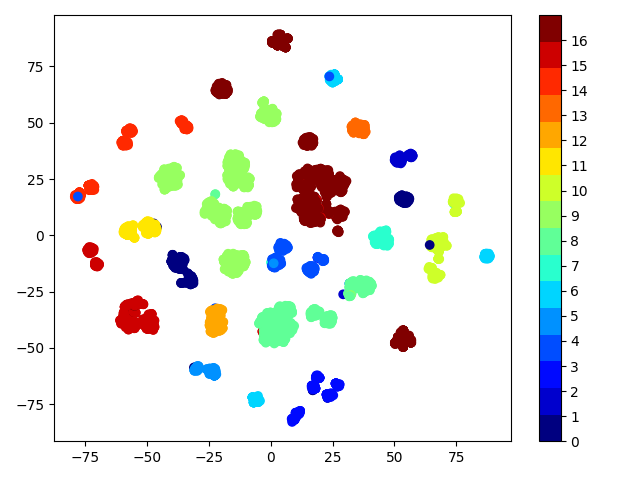}
    \caption{Random target model}
    \label{fig:target_features_random}
  \end{subfigure}
  \begin{subfigure}{0.5\textwidth}
    \centering
    \includegraphics[width=7cm]{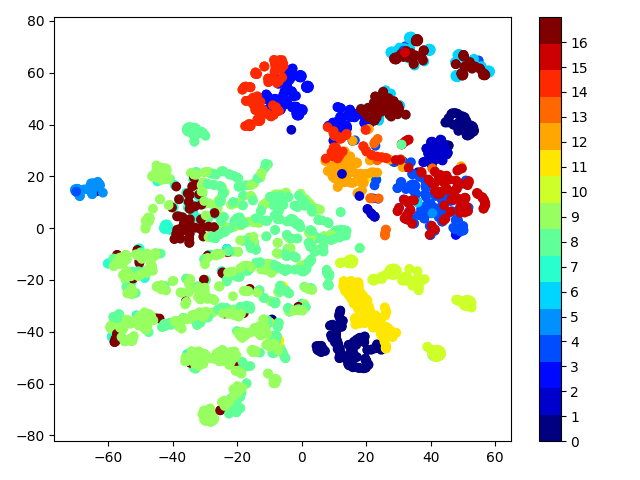}
    \caption{Trained target model}
    \label{fig:trained_features_random}
  \end{subfigure}  
\caption{The t-SNE projected feature representations of the target model in Montezuma's Revenge task. The colors correspond to different rooms.}
\label{fig:cnd_feature_space}
\end{figure}

\begin{figure}[t!]
  \centering
  \begin{subfigure}[b]{0.3\textwidth}
    \centering
    \includegraphics[width=4cm]{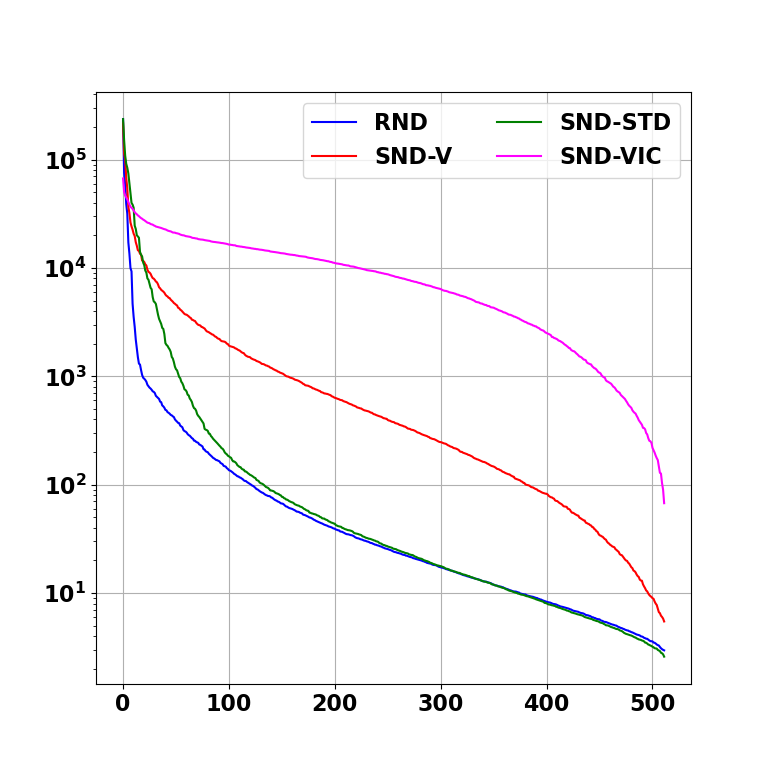}
    \caption{Montezuma's Revenge}
    \label{fig:eigen2a}
  \end{subfigure}
  \begin{subfigure}[b]{0.3\textwidth}
    \centering
    \includegraphics[width=4cm]{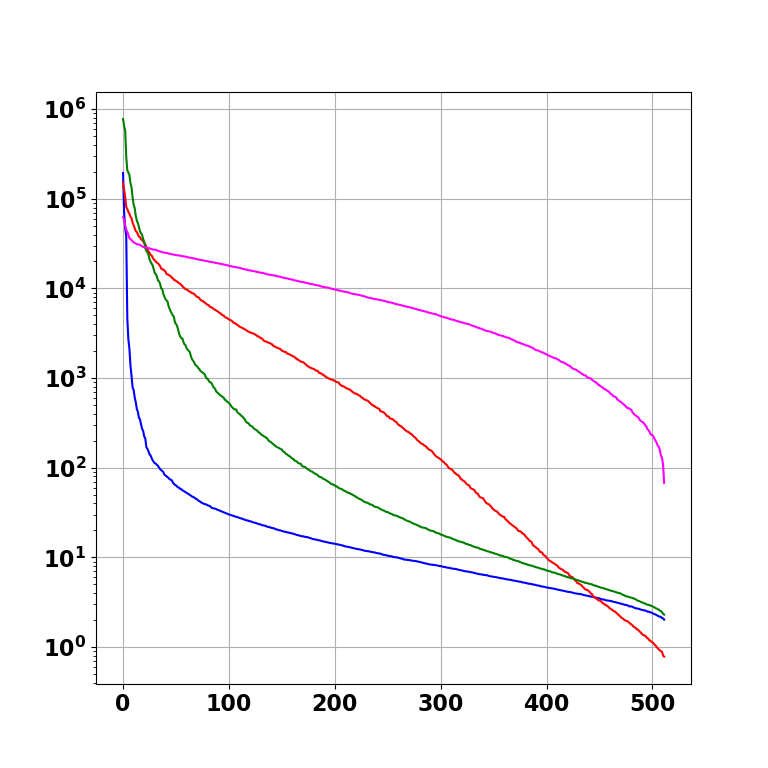}
    \caption{Gravitar}
    \label{fig:eigen2b}
  \end{subfigure}  
  \begin{subfigure}[b]{0.3\textwidth}
    \centering
    \includegraphics[width=4cm]{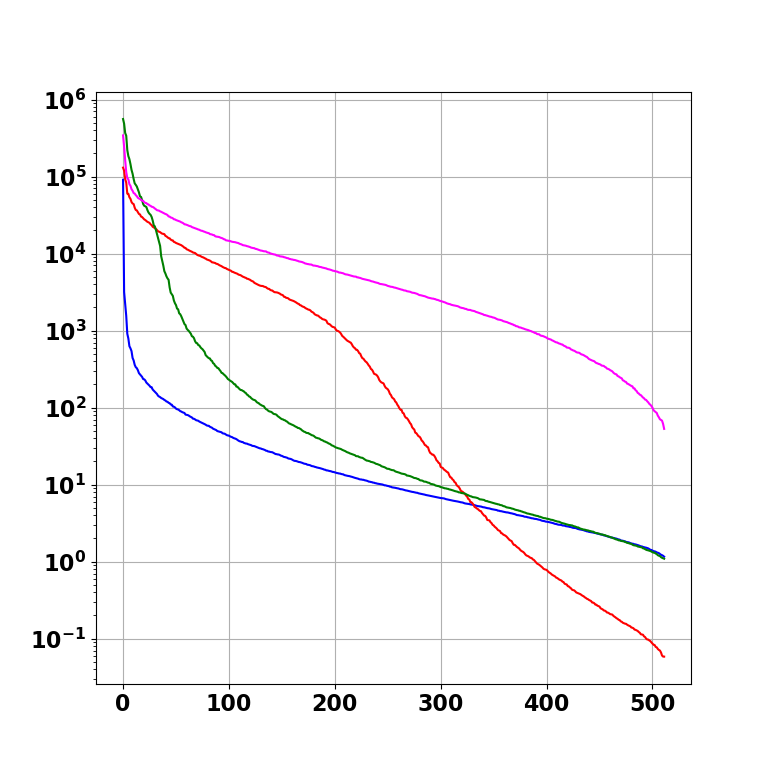}
    \caption{Venture}
    \label{fig:eigen2c}
  \end{subfigure}
  \begin{subfigure}[b]{0.3\textwidth}
    \centering
    \includegraphics[width=4cm]{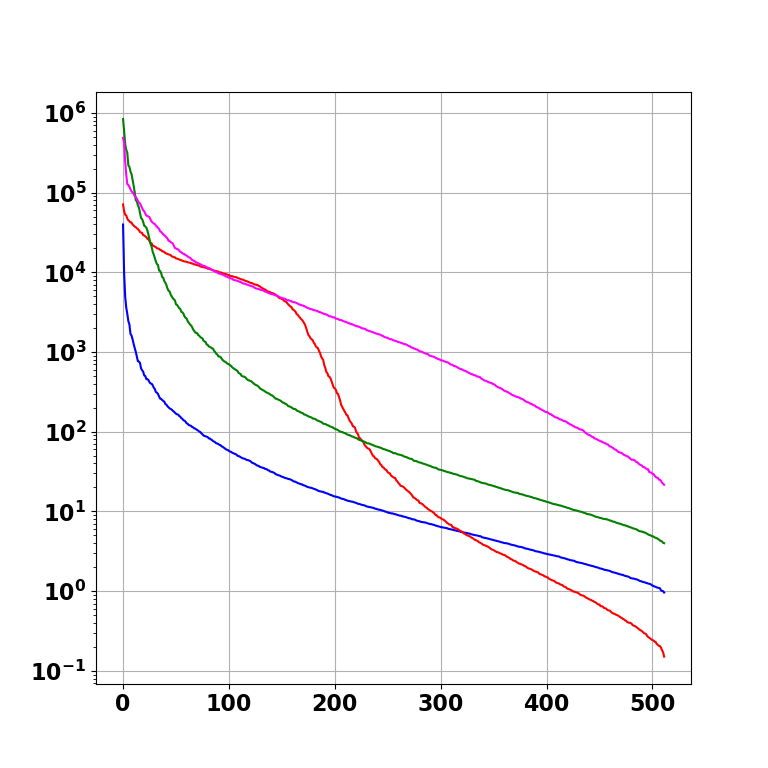}
    \caption{Private Eye}
    \label{fig:eigen2d}
  \end{subfigure} 
  \begin{subfigure}[b]{0.3\textwidth}
    \centering
    \includegraphics[width=4cm]{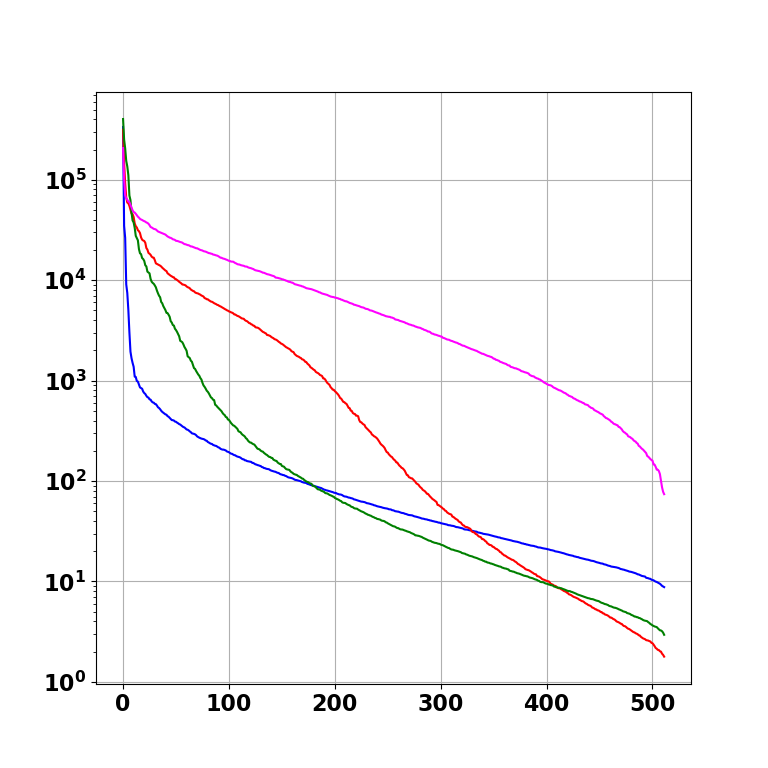}
    \caption{Solaris}
    \label{fig:eigen2e}   
  \end{subfigure}
    \begin{subfigure}[b]{0.3\textwidth}
    \centering
    \includegraphics[width=4cm]{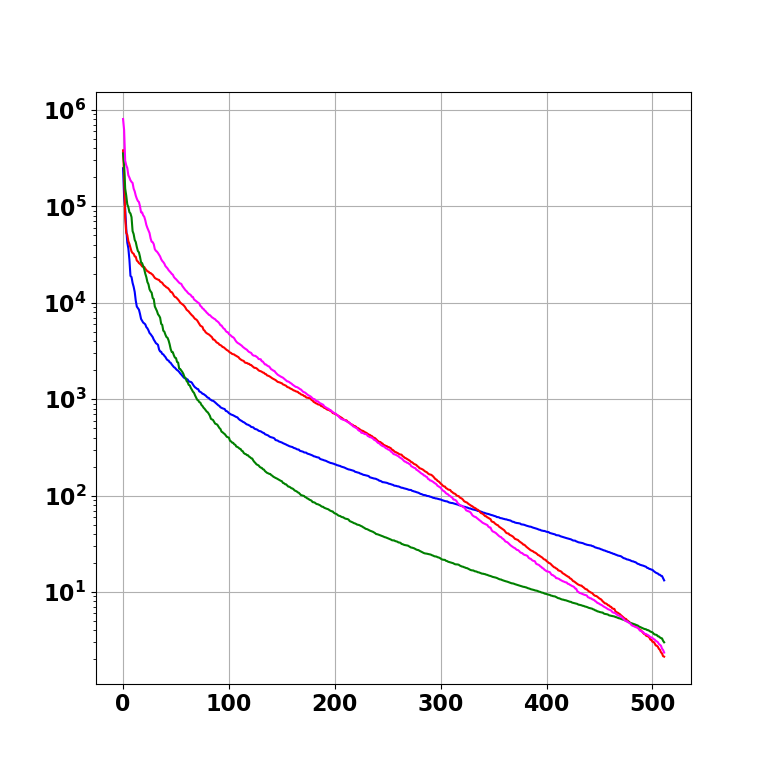}
    \caption{Caveflyer}
    \label{fig:eigen2f}
  \end{subfigure}
  \begin{subfigure}[b]{0.3\textwidth}
    \centering
    \includegraphics[width=4cm]{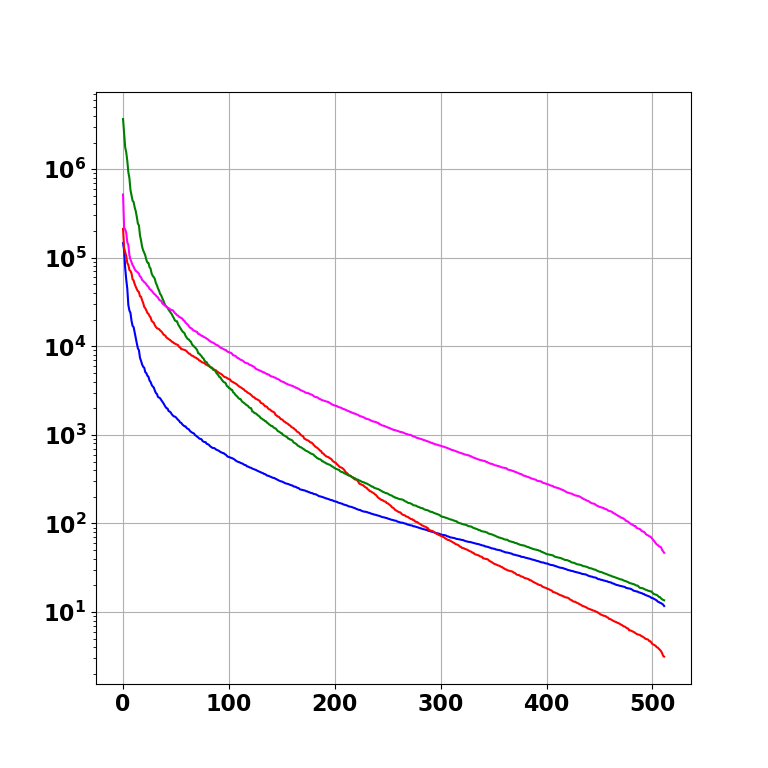}
    \caption{Coinrun}
    \label{fig:eigen2g}
  \end{subfigure}
  \begin{subfigure}[b]{0.3\textwidth}
    \centering
    \includegraphics[width=4cm]{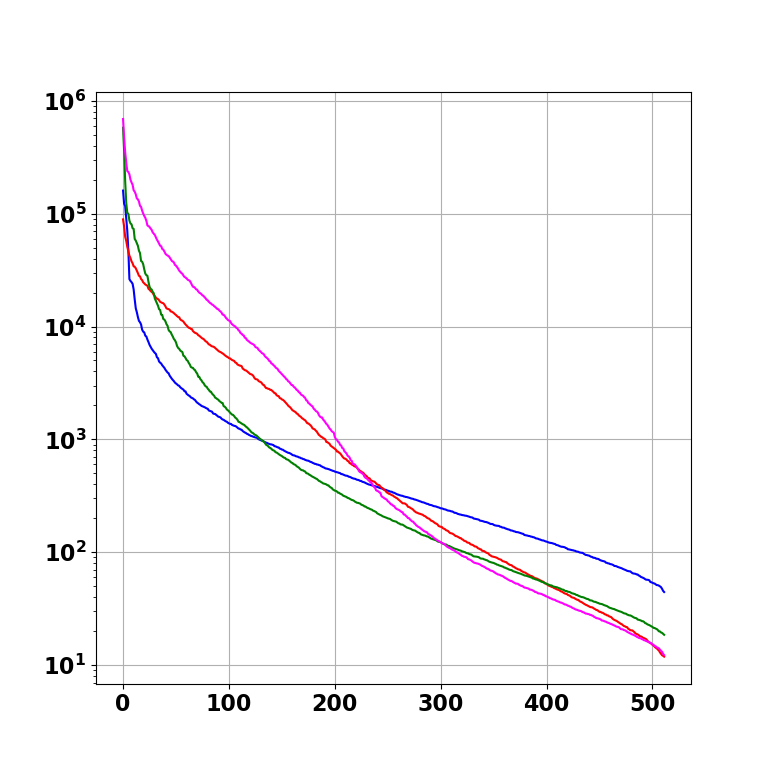}
    \caption{Jumper}
    \label{fig:eigen2h}
  \end{subfigure}
    \begin{subfigure}[b]{0.3\textwidth}
    \centering
    \includegraphics[width=4cm]{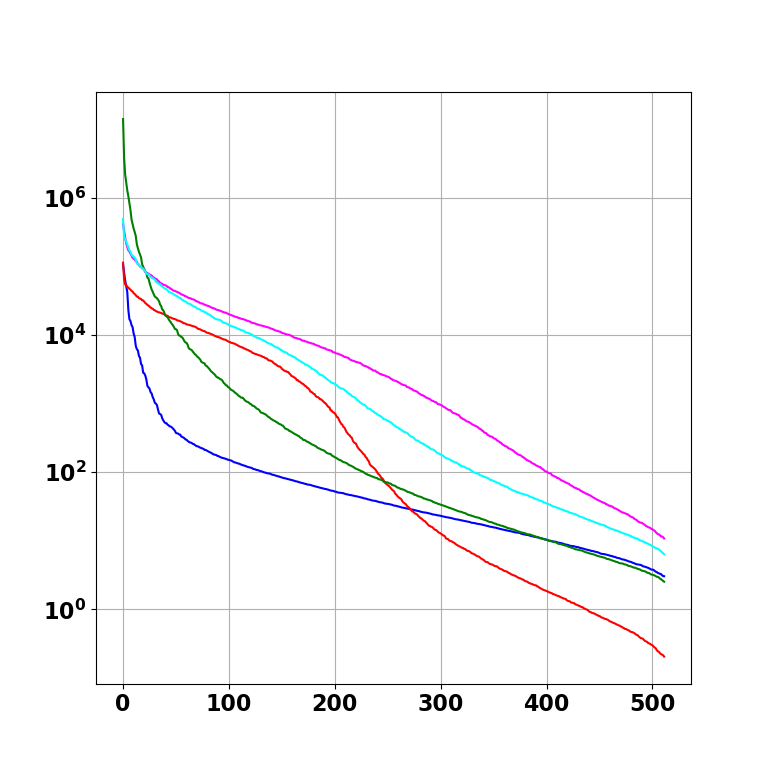}
    \caption{Climber}
    \label{fig:eigen2i}
  \end{subfigure}
\caption{Descendingly ordered eigenvalues of a linear envelope obtained using the PCA method, which show the stretching of the feature space in individual dimensions. 
The horizontal axis denotes the indices of eigenvalues, the vertical axis shows the magnitude of eigenvalue on a logarithmic scale.
Based on these data, we aimed to find out if there is a connection between the shape of the feature space and the performance of the given model. The graph for Pitfall was omitted since it looked very similar to Private Eye. }
\label{fig:cnd_eigen}
\end{figure}

\begin{table}[t!]
\scriptsize
\centering
\caption{Description of the target model feature space created by four selected methods. For the evaluation, we used the following parameters: mean value and standard deviation of the $L_{2}$-norm of features, 25th, 50th, 75th, and 95th percentiles of eigenvalues of a linear envelop to obtain a rough representation of stretching of the feature space in individual dimensions. To this, we add the maximum achieved external reward ($\max (r_{\rm ext})$), so that it is possible to search for a connection between the parameters of the feature space and the performance of the method.}
\begin{tabular}{l|l|cccccc}
\hline
 Environment & Method & $\max({r_{\rm ext}})$ & $L_2$-norm & $Q_{25}$ & $Q_{50}$ & $Q_{75}$ & $Q_{95}$ \\
\hline\hline
\multirow{4}{*}{Montezuma}
& \multicolumn{1}{l|}{RND} & \multicolumn{1}{c}{9} & \multicolumn{1}{c}{1.93 $\pm$ 1.15} & \multicolumn{1}{c}{9} & \multicolumn{1}{c}{24} & \multicolumn{1}{c}{89} & \multicolumn{1}{c}{778} \\
& \multicolumn{1}{l|}{SND-STD} & \multicolumn{1}{c}{16} & \multicolumn{1}{c}{4.85 $\pm$ 2.45} & \multicolumn{1}{c}{9} & \multicolumn{1}{c}{26} & \multicolumn{1}{c}{109} & \multicolumn{1}{c}{6835} \\
& \multicolumn{1}{l|}{SND-VIC} & \multicolumn{1}{c}{17} & \multicolumn{1}{c}{6.22 $\pm$ 2.99} & \multicolumn{1}{c}{3066} & \multicolumn{1}{c}{8429} & \multicolumn{1}{c}{14858} & \multicolumn{1}{c}{25634} \\
& \multicolumn{1}{l|}{SND-V} & \multicolumn{1}{c}{34} & \multicolumn{1}{c}{6.41 $\pm$ 3.96} & \multicolumn{1}{c}{98} & \multicolumn{1}{c}{378} & \multicolumn{1}{c}{1367} & \multicolumn{1}{c}{8944} \\
\hline%\hline
\multirow{4}{*}{Gravitar}
& \multicolumn{1}{l|}{SND-V} & \multicolumn{1}{c}{9} & \multicolumn{1}{c}{11.94 $\pm$ 3.81} & \multicolumn{1}{c}{16} & \multicolumn{1}{c}{341} & \multicolumn{1}{c}{2919} & \multicolumn{1}{c}{24291} \\
& \multicolumn{1}{l|}{SND-STD} & \multicolumn{1}{c}{15} & \multicolumn{1}{c}{11.66 $\pm$ 2.82} & \multicolumn{1}{c}{8} & \multicolumn{1}{c}{30} & \multicolumn{1}{c}{250} & \multicolumn{1}{c}{20402} \\
& \multicolumn{1}{l|}{RND} & \multicolumn{1}{c}{20} & \multicolumn{1}{c}{1.12 $\pm$ 0.81} & \multicolumn{1}{c}{5} & \multicolumn{1}{c}{10} & \multicolumn{1}{c}{24} & \multicolumn{1}{c}{139} \\
& \multicolumn{1}{l|}{SND-VIC} & \multicolumn{1}{c}{21} & \multicolumn{1}{c}{7.26 $\pm$ 3.01} & \multicolumn{1}{c}{2263} & \multicolumn{1}{c}{6837} & \multicolumn{1}{c}{15173} & \multicolumn{1}{c}{27908} \\
\hline%\hline
\multirow{4}{*}{Private Eye}
& \multicolumn{1}{l|}{SND-V} & \multicolumn{1}{c}{7} & \multicolumn{1}{c}{15.47 $\pm$ 6.50} & \multicolumn{1}{c}{2} & \multicolumn{1}{c}{27} & \multicolumn{1}{c}{6801} & \multicolumn{1}{c}{23823} \\
& \multicolumn{1}{l|}{RND} & \multicolumn{1}{c}{9} & \multicolumn{1}{c}{1.40 $\pm$ 0.65} & \multicolumn{1}{c}{3} & \multicolumn{1}{c}{9} & \multicolumn{1}{c}{37} & \multicolumn{1}{c}{410} \\ 
& \multicolumn{1}{l|}{SND-STD} & \multicolumn{1}{c}{9} & \multicolumn{1}{c}{8.03 $\pm$ 4.46} & \multicolumn{1}{c}{15} & \multicolumn{1}{c}{54} & \multicolumn{1}{c}{371} & \multicolumn{1}{c}{24025} \\
& \multicolumn{1}{l|}{SND-VIC} & \multicolumn{1}{c}{10} & \multicolumn{1}{c}{4.29 $\pm$ 3.56} & \multicolumn{1}{c}{232} & \multicolumn{1}{c}{1401} & \multicolumn{1}{c}{6197} & \multicolumn{1}{c}{47342} \\
\hline%\hline
\multirow{4}{*}{Pitfall}
& \multicolumn{1}{l|}{RND} & \multicolumn{1}{c}{0} & \multicolumn{1}{c}{1.31 $\pm$ 0.39} & \multicolumn{1}{c}{3} & \multicolumn{1}{c}{10} & \multicolumn{1}{c}{47} & \multicolumn{1}{c}{410} \\
& \multicolumn{1}{l|}{SND-V} & \multicolumn{1}{c}{0} & \multicolumn{1}{c}{2.81 $\pm$ 1.43} & \multicolumn{1}{c}{2} & \multicolumn{1}{c}{124} & \multicolumn{1}{c}{7358} & \multicolumn{1}{c}{27402} \\
& \multicolumn{1}{l|}{SND-STD} & \multicolumn{1}{c}{0} & \multicolumn{1}{c}{10.64 $\pm$ 2.81} & \multicolumn{1}{c}{8} & \multicolumn{1}{c}{32} & \multicolumn{1}{c}{319} & \multicolumn{1}{c}{39904} \\
& \multicolumn{1}{l|}{SND-VIC} & \multicolumn{1}{c}{0} & \multicolumn{1}{c}{5.62 $\pm$ 1.79} & \multicolumn{1}{c}{153} & \multicolumn{1}{c}{1542} & \multicolumn{1}{c}{9827} & \multicolumn{1}{c}{44126} \\
\hline%\hline
\multirow{4}{*}{Venture}
& \multicolumn{1}{l|}{SND-V} & \multicolumn{1}{c}{14} & \multicolumn{1}{c}{3.67 $\pm$ 3.07} & \multicolumn{1}{c}{1} & \multicolumn{1}{c}{130} & \multicolumn{1}{c}{3921} & \multicolumn{1}{c}{24443} \\
& \multicolumn{1}{l|}{RND} & \multicolumn{1}{c}{18} & \multicolumn{1}{c}{0.68 $\pm$ 0.80} & \multicolumn{1}{c}{4} & \multicolumn{1}{c}{9} & \multicolumn{1}{c}{30} & \multicolumn{1}{c}{188} \\ 
& \multicolumn{1}{l|}{SND-STD} & \multicolumn{1}{c}{18} & \multicolumn{1}{c}{4.50 $\pm$ 3.79} & \multicolumn{1}{c}{4} & \multicolumn{1}{c}{15} & \multicolumn{1}{c}{115} & \multicolumn{1}{c}{32272} \\
& \multicolumn{1}{l|}{SND-VIC} & \multicolumn{1}{c}{18} & \multicolumn{1}{c}{4.91 $\pm$ 3.89} & \multicolumn{1}{c}{1006} & \multicolumn{1}{c}{3644} & \multicolumn{1}{c}{11256} & \multicolumn{1}{c}{41899} \\
\hline%\hline
\multirow{4}{*}{Solaris}
& \multicolumn{1}{l|}{RND} & \multicolumn{1}{c}{55} & \multicolumn{1}{c}{3.68 $\pm$ 3.42} & \multicolumn{1}{c}{29} & \multicolumn{1}{c}{62} & \multicolumn{1}{c}{172} & \multicolumn{1}{c}{776} \\
& \multicolumn{1}{l|}{SND-V} & \multicolumn{1}{c}{65} & \multicolumn{1}{c}{10.19 $\pm$ 6.39} & \multicolumn{1}{c}{13} & \multicolumn{1}{c}{165} & \multicolumn{1}{c}{3337} & \multicolumn{1}{c}{18085} \\
& \multicolumn{1}{l|}{SND-STD} & \multicolumn{1}{c}{81} & \multicolumn{1}{c}{5.77 $\pm$ 4.28} & \multicolumn{1}{c}{12} & \multicolumn{1}{c}{37} & \multicolumn{1}{c}{211} & \multicolumn{1}{c}{11171} \\
& \multicolumn{1}{l|}{SND-VIC} & \multicolumn{1}{c}{87} & \multicolumn{1}{c}{6.00 $\pm$ 4.14} & \multicolumn{1}{c}{1649} & \multicolumn{1}{c}{5079} & \multicolumn{1}{c}{13163} & \multicolumn{1}{c}{32377} \\
\hline\hline
\multirow{4}{*}{Caveflyer}
& \multicolumn{1}{l|}{RND} & \multicolumn{1}{c}{10} & \multicolumn{1}{c}{5.06 $\pm$ 2.98} & \multicolumn{1}{c}{48} & \multicolumn{1}{c}{128} & \multicolumn{1}{c}{474} & \multicolumn{1}{c}{4792} \\
& \multicolumn{1}{l|}{SND-V} & \multicolumn{1}{c}{16} & \multicolumn{1}{c}{10.35 $\pm$ 8.37} & \multicolumn{1}{c}{28} & \multicolumn{1}{c}{293} & \multicolumn{1}{c}{2007} & \multicolumn{1}{c}{20416} \\
& \multicolumn{1}{l|}{SND-STD} & \multicolumn{1}{c}{16} & \multicolumn{1}{c}{9.69 $\pm$ 5.90} & \multicolumn{1}{c}{11} & \multicolumn{1}{c}{34} & \multicolumn{1}{c}{206} & \multicolumn{1}{c}{13452} \\
& \multicolumn{1}{l|}{SND-VIC} & \multicolumn{1}{c}{16} & \multicolumn{1}{c}{9.89 $\pm$ 5.70} & \multicolumn{1}{c}{23} & \multicolumn{1}{c}{271} & \multicolumn{1}{c}{2638} & \multicolumn{1}{c}{49517} \\
\hline%\hline
\multirow{4}{*}{Climber}
& \multicolumn{1}{l|}{RND} & \multicolumn{1}{c}{0} & \multicolumn{1}{c}{7.47 $\pm$ 3.53} & \multicolumn{1}{c}{12} & \multicolumn{1}{c}{32} & \multicolumn{1}{c}{106} & \multicolumn{1}{c}{1500} \\
& \multicolumn{1}{l|}{SND-V} & \multicolumn{1}{c}{11} & \multicolumn{1}{c}{17.31 $\pm$ 6.45} & \multicolumn{1}{c}{2} & \multicolumn{1}{c}{53} & \multicolumn{1}{c}{5091} & \multicolumn{1}{c}{25463} \\
& \multicolumn{1}{l|}{SND-STD} & \multicolumn{1}{c}{11} & \multicolumn{1}{c}{48.56 $\pm$ 22.16} & \multicolumn{1}{c}{12} & \multicolumn{1}{c}{63} & \multicolumn{1}{c}{788} & \multicolumn{1}{c}{49506} \\
& \multicolumn{1}{l|}{SND-VIC} & \multicolumn{1}{c}{11} & \multicolumn{1}{c}{13.41 $\pm$ 4.01} & \multicolumn{1}{c}{144} & \multicolumn{1}{c}{2313} & \multicolumn{1}{c}{14132} & \multicolumn{1}{c}{78074} \\
\hline%\hline
\multirow{4}{*}{Coinrun}
& \multicolumn{1}{l|}{RND} & \multicolumn{1}{c}{10} & \multicolumn{1}{c}{5.07 $\pm$ 3.00} & \multicolumn{1}{c}{40} & \multicolumn{1}{c}{109} & \multicolumn{1}{c}{389} & \multicolumn{1}{c}{4029} \\
& \multicolumn{1}{l|}{SND-V} & \multicolumn{1}{c}{10} & \multicolumn{1}{c}{12.02 $\pm$ 7.20} & \multicolumn{1}{c}{23} & \multicolumn{1}{c}{149} & \multicolumn{1}{c}{2473} & \multicolumn{1}{c}{21553} \\
& \multicolumn{1}{l|}{SND-STD} & \multicolumn{1}{c}{10} & \multicolumn{1}{c}{29.23 $\pm$ 17.06} & \multicolumn{1}{c}{54} & \multicolumn{1}{c}{200} & \multicolumn{1}{c}{1651} & \multicolumn{1}{c}{76730} \\
& \multicolumn{1}{l|}{SND-VIC} & \multicolumn{1}{c}{10} & \multicolumn{1}{c}{11.23 $\pm$ 5.77} & \multicolumn{1}{c}{332} & \multicolumn{1}{c}{1140} & \multicolumn{1}{c}{5414} & \multicolumn{1}{c}{44014} \\
\hline%\hline
\multirow{4}{*}{Jumper}
& \multicolumn{1}{l|}{RND} & \multicolumn{1}{c}{10} & \multicolumn{1}{c}{8.63 $\pm$ 2.19} & \multicolumn{1}{c}{139} & \multicolumn{1}{c}{337} & \multicolumn{1}{c}{1008} & \multicolumn{1}{c}{6733} \\
& \multicolumn{1}{l|}{SND-V} & \multicolumn{1}{c}{10} & \multicolumn{1}{c}{14.97 $\pm$ 5.49} & \multicolumn{1}{c}{63} & \multicolumn{1}{c}{314} & \multicolumn{1}{c}{3310} & \multicolumn{1}{c}{20896} \\
& \multicolumn{1}{l|}{SND-STD} & \multicolumn{1}{c}{10} & \multicolumn{1}{c}{15.51 $\pm$ 3.73} & \multicolumn{1}{c}{61} & \multicolumn{1}{c}{189} & \multicolumn{1}{c}{1044} & \multicolumn{1}{c}{22008} \\
& \multicolumn{1}{l|}{SND-VIC} & \multicolumn{1}{c}{10} & \multicolumn{1}{c}{20.79 $\pm$ 6.18} & \multicolumn{1}{c}{47} & \multicolumn{1}{c}{257} & \multicolumn{1}{c}{6271} & \multicolumn{1}{c}{74860} \\
\hline\hline
\end{tabular}
\label{tab:analysis1}
\end{table}

\label{sec:analysis}
\subsection{Analysis of results}

We analyzed the results using three different methods in a reasonable order. 
The first method compares the models from the view of the quality of an intrinsic reward acquired during training. The second approach focuses on
the distribution of the representations in the feature space and shows how the distances of points in the state space are transformed in the feature space. The third approach 
contains the spatial analysis of the feature spaces based on computing eigenvalues of the linear envelope of the feature space.

\subsubsection{Quality of an intrinsic reward}
\label{sec:analysis_1}

By means of the first analysis, we want to demonstrate that the presented results were achieved thanks to a better intrinsic reward signal which is superior to the signal of the RND method. For different regularisation losses, we wanted to understand its time evaluation and show the ability to provide a large IM signal for previously unseen states. For the purpose of exploration, the most important is the ability to detect near-future states that are very close to already seen ones. We collected a set of 2700 states from our best agent playing Montezuma's Revenge. During the experiment, we trained the IM modules only on past data and tested them on future data. The testing batch was selected from the following 4-time horizons, with respect to the agent being in step $n$, and testing batch indices $m$:
\begin{enumerate}
    \item past: already seen states, $m<n$
    \item near future: $n < m < n+128$ steps in the future
    \item far future: $m > n$
    \item random: any batch from the set
\end{enumerate}

We hypothesized that the Random Network distillation would provide a sufficient signal only at the beginning of learning. Convergence to zero leads to limited exploration abilities. This degradation corresponds to our results in Figure~\ref{fig:nov_rnd_result_summary}.
On the other hand, continuously updated target models can provide useful signals for the entire run. The corresponding results are displayed in Figures \ref{fig:nov_nce_result_summary}--d. %\ref{fig:nov_mse_result_summary}, and \ref{fig:nov_vicreg_result_summary}. 
On all three losses, the intrinsic motivation is much higher for unseen states, and not converging into zero. The strong peaks for the near future correspond to finding a new room. The self-supervised regularisation prevents collapsing the motivation signal to zero. 
This insight implies requirements for an exploration signal.
For future research, this also provides a simple methodology for testing exploration abilities, without training the whole RL agent, which can be time-consuming. 

\subsubsection{Distribution of the representations in the feature spaces}
\label{sec:analysis_2}

In the second analysis, we investigated how the distances of points in the state space are mapped to the feature space. For a better idea of how the SND methods work, we visualized the feature space by collecting 1000 states in Montezuma's Revenge environment from a trained agent and calculating their distance matrix. Figure~\ref{fig:snd_state_features} illustrates how much of the structure of the state space is preserved and transferred to the feature space. The patterns on the diagonal correspond to individual rooms in the environment where the states are similar to each other and therefore their mutual distance is small (blue color), while the distances to the states from other rooms are large (red color). One can notice that in the case of both RND and SND-STD, a similar structure of distances is preserved as in the case of states, but with smaller differences. When comparing RND and SND-STD with each other, it can be seen that the feature vectors created by the SND-STD method still have larger distances, which is indicated by the light blue color. In contrast, the SND-V method did not preserve the original structure of the state space, although some fragments can be seen there that can be mapped to the input data. In any case, it can be seen that initially, similar states are significantly further apart in this space. Finally, SND-VIC is extremely contrastive and apart from the main diagonal, there is no distinct original structure. However, the light red color indicates that the distances for almost any state are larger than the three previous methods, which indicates better discriminative capabilities of SND-VIC. The enhanced discriminative capabilities of SND methods can explain more efficient novelty detection that provides a better intrinsic reward signal.

To demonstrate the described property, we can visualize the learned feature vectors $Z$ in 2D using the t-SNE method \citep{tSNE2008}. Figure~\ref{fig:cnd_feature_space} shows the resulting features of the trained target model on Atari Montezuma's Revenge task. The randomly initialized trained network (the same as in \cite{burda2018exploration}, Figure~\ref{fig:target_features_random}) can differentiate well between different rooms, however within the room the variance is low, pointing to the lack of exploration abilities. On the other hand, the self-supervised regularized target model in Figure~\ref{fig:cnd_feature_space} provides a much larger variance of features, which provides a more sensitive novelty detection signal.

\subsubsection{Spatial analysis of the feature spaces}
\label{sec:analysis_3}

The main goal of the third analysis was to find out the differences (not only visually) among the individual spaces of features, to describe them with some quantities, and to find a possible connection to the performance of the algorithm and the mentioned quantities. From the set of examined models for one environment, we always selected the model with the highest obtained reward and generated 10,000 samples of input states by running it in the environment. Subsequently, each model generated feature vectors for a sample of previously collected input states. Hence, we obtained feature space samples $Z$ of each model. Using principal component analysis (PCA) we found the linear envelope of the high-dimensional manifold that forms the feature space. We examined the mean value and especially the variance of the feature vectors, and also the eigenvalues using PCA, which indicate the basic shape of the feature space (i.e. the sizes of the individual dimensions).

The results of this analysis are shown in Figure~\ref{fig:cnd_eigen} and Table~\ref{tab:analysis1}. For the evaluation, we decided to use the following parameters: mean value and standard deviation of the $L_{2}$-norm of features, 25th, 50th, 75th and 95th percentiles of eigenvalues to obtain a rough representation of stretching of the feature space in individual dimensions.
It can be seen that in almost all cases the RND target model has smaller eigenvalues than the SND models. This is also evident for the $L_2$-norm values that the entire RND feature space has a smaller volume compared to the SND feature space. 
On the other hand, RND and SND-STD are similar in shape. Their curves have a convex shape, with SND-STD having more stretched dimensions. SND-V and SND-VIC also have similarly concave shapes, but SND-V stretches only about half of the available dimensions and then usually falls more steeply. 
The shape of the SND-VIC curve is ensured by the variance (Eq.~\ref{eq:sndvic1}) and covariance (Eq.~\ref{eq:sndvic2}) components of its loss function. The missing decorrelation term in the SND-V loss function (Eq.~\ref{eq:sndv2}) results in an uneven stretching of the dimensions.

In contrast, the shape of the SND-STD curve is convex and the dimensions are used unevenly. After these analyses, we tried to improve the variance within the dimensions by adding a regularization term (Eq.~\ref{eq:sndstd5}) which tried to maximize the variance within the feature vector. However, such a term had an expansive effect on the feature space and it was not possible to give it much weight, because the loss function (Eq.~\ref{eq:sndstd3}) of the ST-DIM algorithm has an expansive effect, and the addition of another expansive term led to problems with the uncontrolled expansion of the feature space. 
Despite the small influence of the variance component of the loss function, the performance of SND-STD improved and it helped prevent agents from getting stuck in certain cases. If we compare RND and SND-STD feature spaces (in terms of eigenvalues) they look similar, but the latter model was able to achieve better results in 7 out of 9 environments. Our findings show that when training the target model, it is important to enforce the decorrelation of features and the equal use of all dimensions of the feature space. Such a model seems to be relatively robust and sufficiently sensitive to novelty.

Interestingly, for the Pitfall task (not shown in Figure~\ref{fig:cnd_eigen}), despite their failure, our methods still tried to take advantage of the feature space dimensions. 
From the analysis of the trained agents, we saw that they were able to explore several rooms, but in each, there were enough moving objects that made the given state space rich and thus made it difficult to train the predictor model. This led to a very slow decrease in the intrinsic reward (we observed a similar behavior after short training sessions in other environments). We assume that with a larger number of training steps, the agent would eventually be able to reach the reward.

%%%%%%%%%%%%%%%%%%%%%%%%%%%%%%%%%%%%%%%%%%%%%%%%%%%%%%%%%%%%%%%%%%%%%%%% CHAPTER 5
\section{Discussion}
\label{sec:discussion}

At the beginning, we asked ourselves the following questions: Are the features provided by the original RND target model sufficient? Can the model learn better features? In the paper Random Network Distillation \citep{burda2018exploration}, the orthogonal weight initialization method was used to set the parameters of the target model and distribute its representations over the state space, which provides a sustainable intrinsic reward. We assumed that if it is possible to find suitable algorithms for training the target model, it will also be possible to improve the results of the RND model. Based on these considerations, we proposed a number of SND methods.

The methods worked very well in most cases, so in the analysis, we focused on their deeper understanding. First, we used a geometric approach trying to capture, at least in rough outlines, the properties of feature spaces. A comparison between a randomly initialized feature space and a feature space formed using one of the SND algorithms supports the correctness of our assumptions that self-supervised algorithms can distinguish even subtle differences within the state space. This turned out to be one of the weaknesses of the RND algorithm which, while being good at differentiating between sufficiently different states (e.g.~different rooms in Montezuma's Revenge), was placing similar states close to each other in the feature space, making the work of the predictor model easier. In the experiments, we thus observed a decrease in the standard deviation of the average intrinsic reward per episode, which meant that most of the visited states generated a similar reward.

Last but not least, we performed an analysis of novelty detection abilities of selected methods. After comparison with RND as a baseline, we can conclude that this baseline suffers from an IM-based reward vanishing problem. After adding the regularisation to the target model, much better features were obtained with a significant change compared to the baseline. Intrinsic reward vanishing disappears for all the tested losses. This is cross-validated also on t-SNE features visualization, where regularised features yield much higher variance, which means a larger sensitivity to novelty.

We experimented with different target model architectures, different augmentations and intrinsic reward scaling. We found that what worked best was the target model using the ELU activation function and only one fully connected layer, and the predictor model having three hidden layers. 
From the tested augmentations (noise, random tile masking, random convectional filter) we found the best performance for a combination of uniform noise with random tiles masking. As a questionable augmentation remains random downsampling and upsampling back, which could help remove noise while preserving representative information in the state vectors. We suggest investigating this idea for the next research.
The scaling of an intrinsic reward reveals a large sensitivity of the model to this hyperparameter. The best working value was 0.5, but we think that this value should be optimized separately for each specific environment.  

We did not specifically investigate the robustness of SND methods concerning the initialization of the target model (which was again a problem with RND). We assume that self-supervised learning algorithms can cope with a poorly initialized model to a certain extent, but from the training experience, we found that it is better to initialize the target models of SND-STD and SND-VIC to small values ($gain = 0.5$) and let it expand itself while SND-V was initialized like RND to higher values ($gain = \sqrt{2}$).

Our experiments have shown that if the ST-DIM algorithm works on an incomplete dataset that takes on new samples (the authors probably did not test it in such conditions), there is instability and an exponential increase of activity in the feature space at certain moments. This is related to the use of a cross-entropy loss function in its core (which does not limit the values of inputs -- logits) where derivatives can reach large values and subsequently inflate the entire feature space. 

During the development of the model, we found that it is best to minimize the $L_2$-norm of logits that enter the cross-entropy. We also tried to maximize the entropy of the distributions generating the respective logits and minimize the $L_2$-norm of global features, but both approaches failed to sufficiently stabilize the algorithm.
As another thing, we investigated the effect of state preprocessing on the performance of the SND-STD model, but it had no significant effect on the agent performance.

A direct extension of SND methods will be the merging of the target model with the model to which the actor and critic are connected. We have already done some pilot research in this direction and it seems to be a feasible task. This would greatly optimize the entire model and speed up its training in terms of computing time.

The SND methods provide room for experimenting with other suitable algorithms for training the target model. Since none of our three tested methods proved universally superior (i.e. that it would beat the others in all environments), it can be concluded that different environments require slightly different representations of the target model. On the positive side, we first optimized these methods in the Atari environments and were then able to port them to ProcGen with minimal fine-tuning, immediately achieving very good results.

At the same time, we think that this approach can be an inspiration for a new class of algorithms that will specialize in creating feature mapping capturing the relationship of the environment to the agent itself since current self-supervised methods are agnostic to these relationships.

%%%%%%%%%%%%%%%%%%%%%%%%%%%%%%%%%%%%%%%%%%%%%%%%%%%%%%%%%%%%%%%%%%%%%%%% CHAPTER 6
\section{Conclusion}
\label{sec:conclusion}

In this work, we presented novel methods based on Self-Supervised Network Distillation (SND) that can help solve tasks in sparse reward environments using an intrinsic motivation signal.
The main idea of SND is to distill knowledge from the target model using the predictor model, while the target model builds its state representations using self-supervised algorithms. 
The difference between the representations of the target model and the predictor model serves as a motivational signal. We adapted three existing self-supervised algorithms for training the target model and tested them experimentally on a set of environments that are considered difficult to explore.
The SND methods are a generalization of the RND method based on the novelty detection approach
where the target model is randomized and fixed.
Three analytical methods we used revealed the weaknesses of the RND approach -- the need for good initialization, the low variance of intrinsic reward in different states, and the loss of the motivational signal due to the adaptation of the predictor model.
Through empirical experiments, we were able to demonstrate the superior performance of the SND methods over our baseline and RND agents in different game environments.
This highlights the effectiveness and versatility of the SND methods for improving learning in sparse reward scenarios and demonstrates their potential as a valuable tool for reinforcement learning applications.

\section*{Acknowledgement}

\noindent
The authors thank prof.~Peter Ti\v{n}o from the University of Birmingham, UK, for careful reading of the manuscript and constructive feedback. The third author was supported in part by the Slovak Grant Agency for Science (VEGA), project 1/0373/23.

%%%%%%%%%%%%%%%%%%%%%%%%%%%%%%%%%%%%%%%%%%%%% BIBLIOGRAPHY
\vspace{3,16314mm}
% \nocite{*}
\bibliographystyle{apalike}  %preco nie toto?
\bibliography{references}

\end{document}